\documentclass[10pt,twocolumn,letterpaper]{article}

\usepackage[pagenumbers]{cvpr} 

\usepackage{graphicx, amsmath, amssymb, caption, subcaption, multirow, overpic, textpos}
\usepackage[table]{xcolor}
\usepackage[british, english, american]{babel}
\usepackage[pagebackref=false, breaklinks=true, letterpaper=true, colorlinks,
            citecolor=citecolor, linkcolor=linkcolor, bookmarks=false]{hyperref}
\definecolor{citecolor}{HTML}{0071BC}
\definecolor{linkcolor}{HTML}{ED1C24}

\def\cvprPaperID{**}
\def\confName{****}
\def\confYear{****}

\newlength\savewidth\newcommand\shline{\noalign{\global\savewidth\arrayrulewidth
  \global\arrayrulewidth 1pt}\hline\noalign{\global\arrayrulewidth\savewidth}}
\newcommand{\tablestyle}[2]{\setlength{\tabcolsep}{#1}\renewcommand{\arraystretch}{#2}\centering\footnotesize}
\renewcommand{\paragraph}[1]{\vspace{1.25mm}\noindent\textbf{#1}}

\newcolumntype{x}[1]{>{\centering\arraybackslash}p{#1pt}}
\newcolumntype{y}[1]{>{\raggedright\arraybackslash}p{#1pt}}
\newcolumntype{z}[1]{>{\raggedleft\arraybackslash}p{#1pt}}

\newcommand{\app}{\raise.17ex\hbox{$\scriptstyle\sim$}}

\newcommand{\x}{{\times}}
\definecolor{deemph}{gray}{0.6}
\newcommand{\gc}[1]{\textcolor{deemph}{#1}}
\definecolor{baselinecolor}{gray}{.9}
\newcommand{\baseline}[1]{\cellcolor{baselinecolor}{#1}}
\newcommand{\authorskip}{\hspace{2.5mm}}

\begin{document}
\title{
\vspace{-1mm}\Large Masked Autoencoders Are Scalable Vision Learners\vspace{-3mm}}
\author{
 Kaiming He$^{*,\dagger}$ \authorskip Xinlei Chen$^{*}$ \authorskip Saining Xie \authorskip
 Yanghao Li \authorskip Piotr Doll\'ar \authorskip Ross Girshick \\[2mm]
 \small $^{*}$equal technical contribution \qquad $^{\dagger}$project lead \\[2mm]
 Facebook AI Research (FAIR)\vspace{-4mm}
}
\maketitle

\begin{abstract}
This paper shows that masked autoencoders (MAE) are scalable self-supervised learners for computer vision. Our MAE approach is simple: we mask random patches of the input image and reconstruct the missing pixels. It is based on two core designs. First, we develop an \mbox{\emph{asymmetric}} encoder-decoder architecture, with an encoder that operates only on the visible subset of patches (without mask tokens), along with a lightweight decoder that reconstructs the original image from the latent representation and mask tokens. Second, we find that masking a high proportion of the input image, \eg, 75\%, yields a nontrivial and meaningful self-supervisory task. Coupling these two designs enables us to train large models efficiently and effectively: we accelerate training (by 3$\x$ or more) and improve accuracy. Our scalable approach allows for learning high-capacity models that generalize well: \eg, a vanilla \mbox{ViT-Huge} model achieves the best accuracy (87.8\%) among methods that use only ImageNet-1K data. Transfer performance in downstream tasks outperforms supervised pre-training and shows promising scaling behavior.
\end{abstract}

\section{Introduction}
\label{sec:intro}

\begin{figure}[t]\centering
\includegraphics[width=0.98\linewidth]{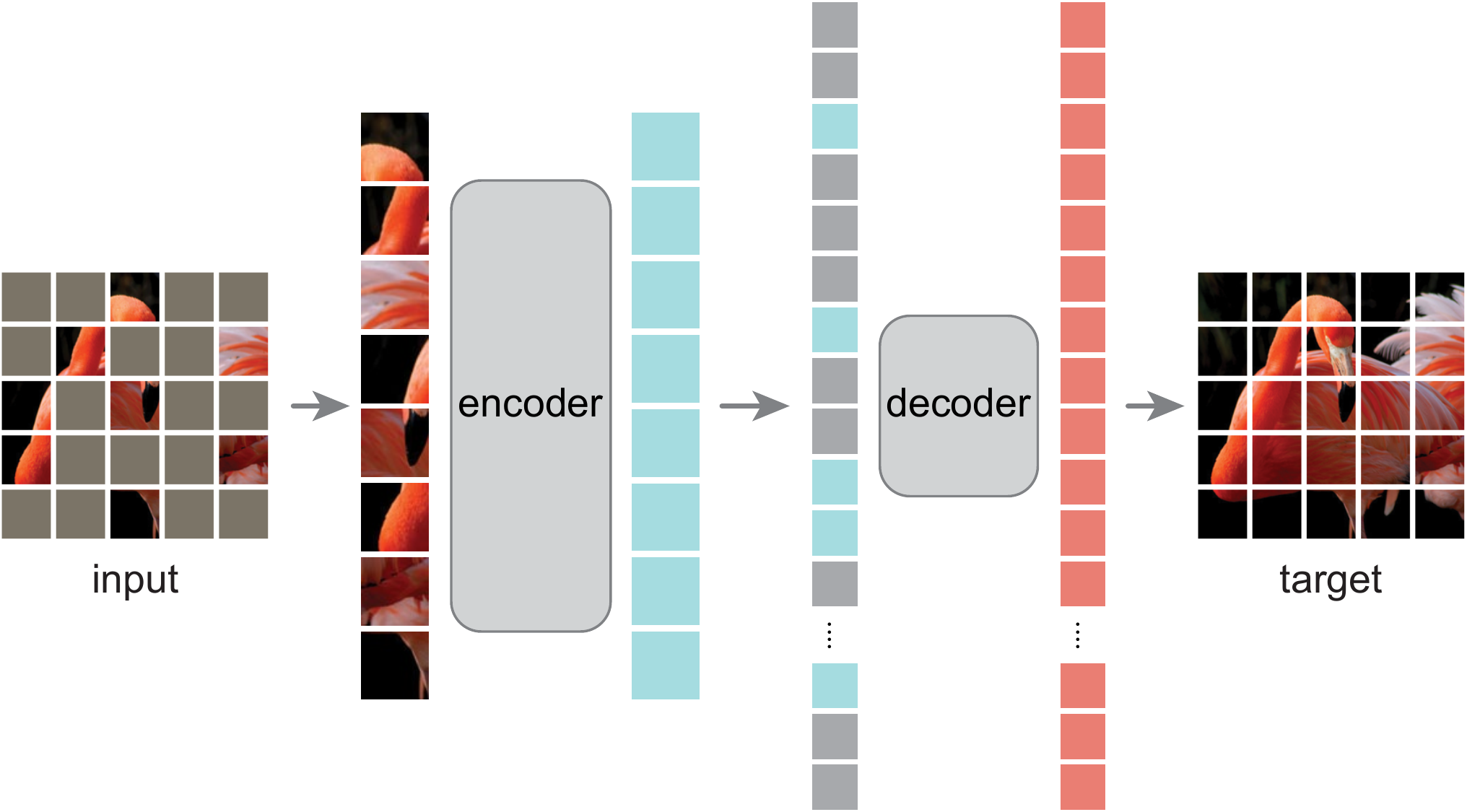}
\caption{\textbf{Our MAE architecture}. During pre-training, a large random subset of image patches (\eg, 75\%) is masked out. The encoder is applied to the small subset of \emph{visible patches}. Mask tokens are introduced \textit{after} the encoder, and the full set of encoded patches and mask tokens is processed by a small decoder that reconstructs the original image in pixels. After pre-training, the decoder is discarded and the encoder is applied to uncorrupted images (full sets of patches) for recognition tasks.}
\label{fig:arch}
\end{figure}

Deep learning has witnessed an explosion of architectures of continuously growing capability and capacity \cite{Krizhevsky2012, He2016, Vaswani2017}. Aided by the rapid gains in hardware, models today can easily overfit one million images \cite{Deng2009} and begin to demand hundreds of millions of---often publicly inaccessible---\textit{labeled} images \cite{Dosovitskiy2021}.

This appetite for data has been successfully addressed in natural language processing (NLP) by self-supervised pre-training. The solutions, based on autoregressive language modeling in GPT \cite{Radford2018, Radford2019, Brown2020} and \emph{masked autoencoding} in BERT \cite{Devlin2019}, are conceptually simple: they remove a portion of the data and learn to predict the removed content. These methods now enable training of generalizable NLP models containing over one hundred billion parameters \cite{Brown2020}.

The idea of masked autoencoders, a form of more general denoising autoencoders \cite{Vincent2008}, is natural and applicable in computer vision as well. Indeed, closely related research in vision \cite{Vincent2010,Pathak2016} preceded BERT. However, despite significant interest in this idea following the success of BERT, progress of autoencoding methods in vision lags behind NLP. We ask: \textit{what makes masked autoencoding different between vision and language}? We attempt to answer this question from the following perspectives:

\begin{figure*}[t]\centering\vspace{-.5em}
\includegraphics[width=0.98\linewidth]{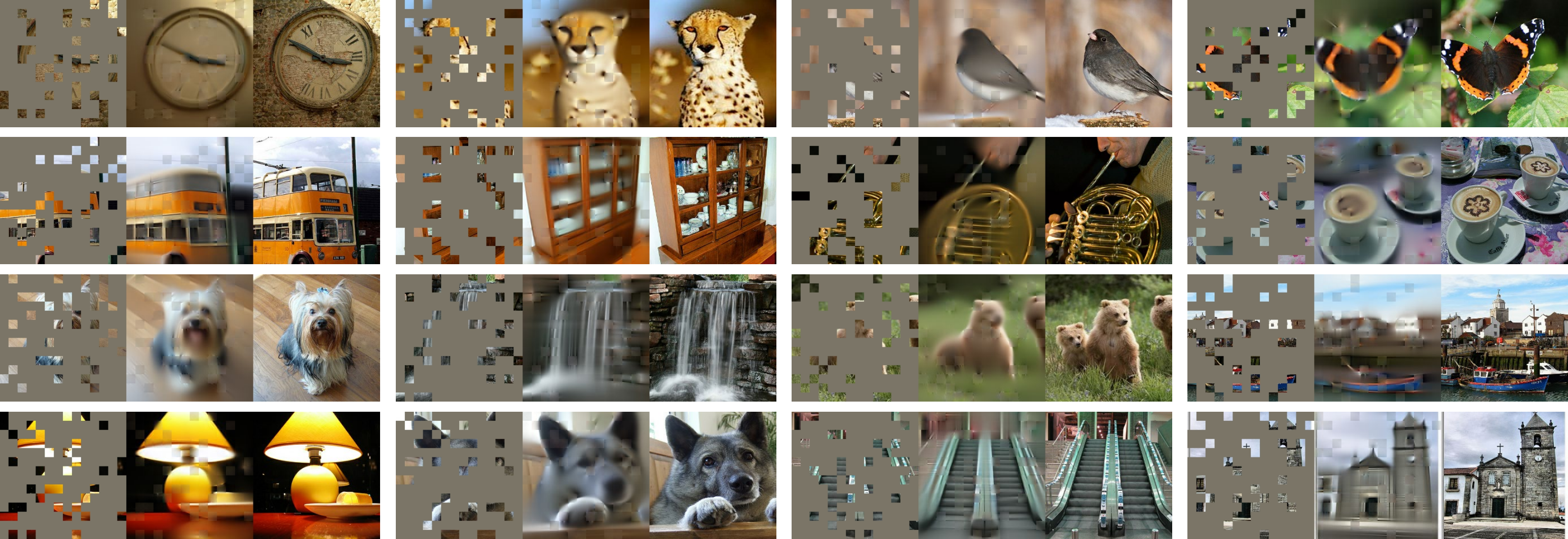}\vspace{-.5em}
\caption{Example results on ImageNet \emph{validation} images. For each triplet, we show the masked image (left), our MAE reconstruction$^\dagger$ (middle), and the ground-truth (right). The masking ratio is {80\%}, leaving only 39 out of 196 patches. More examples are in the appendix.\\ \textit{\footnotesize $^\dagger$As no loss is computed on visible patches, the model output on visible patches is qualitatively worse. One can simply overlay the output with the visible patches to improve visual quality. We intentionally opt not to do this, so we can more comprehensively demonstrate the method's behavior.}}
\label{fig:samples}\vspace{-.5em}
\end{figure*}

\begin{figure*}[t]\centering
\includegraphics[width=0.98\linewidth]{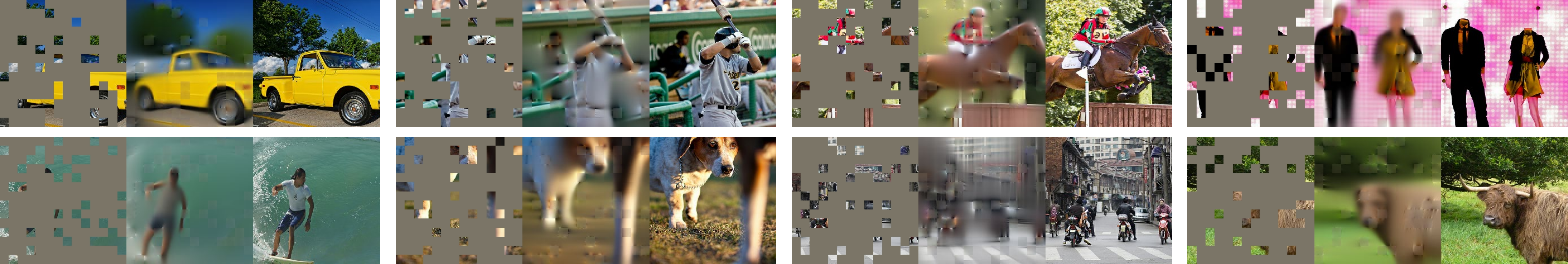}\vspace{-.5em}
\caption{Example results on COCO validation images, using an MAE trained on ImageNet (the same model weights as in Figure~\ref{fig:samples}). Observe the reconstructions on the two right-most examples, which, although different from the ground truth, are semantically plausible.}
\label{fig:samples_coco}\vspace{-2mm}
\end{figure*}

\textbf{(i)} Until recently, architectures were different. In vision, convolutional networks \cite{LeCun1989} were dominant over the last decade \cite{Krizhevsky2012}. Convolutions typically operate on regular grids and it is not straightforward to integrate `indicators' such as mask tokens \cite{Devlin2019} or positional embeddings \cite{Vaswani2017} into convolutional networks. This architectural gap, however, has been addressed with the introduction of Vision Transformers (ViT) \cite{Dosovitskiy2021} and should no longer present an obstacle.

\begin{figure}[t]\centering
\includegraphics[width=0.995\linewidth]{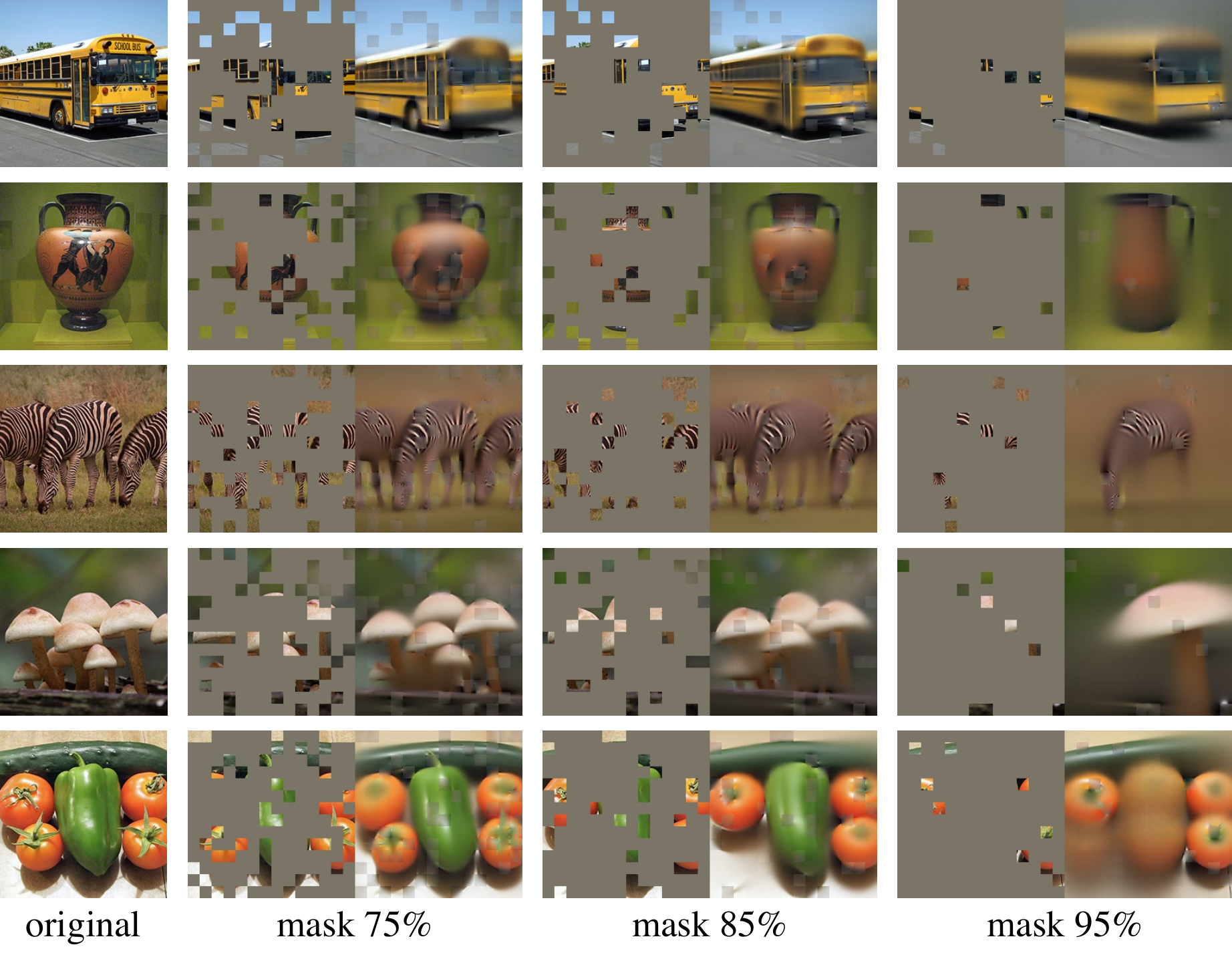}\vspace{-.7em}
\caption{Reconstructions of ImageNet \textit{validation} images using an MAE pre-trained with a masking ratio of 75\% but applied on inputs with higher masking ratios. The predictions differ plausibly from the original images, showing that the method can generalize. }
\label{fig:mask_generalization}
\end{figure}

\textbf{(ii)} Information density is different between language and vision. Languages are human-generated signals that are highly semantic and information-dense. When training a model to predict only a few missing words per sentence, this task appears to induce sophisticated language understanding. Images, on the contrary, are natural signals with heavy spatial redundancy---\eg, a missing patch can be recovered from neighboring patches with little high-level understanding of parts, objects, and scenes. To overcome this difference and encourage learning useful features, we show that a simple strategy works well in computer vision: masking a \textit{very high} portion of random patches. This strategy largely reduces redundancy and creates a challenging self-supervisory task that requires holistic understanding beyond low-level image statistics. To get a qualitative sense of our reconstruction task, see Figures~\ref{fig:samples} -- \ref{fig:mask_generalization}.

\textbf{(iii)} The autoencoder's \textit{decoder}, which maps the latent representation back to the input, plays a different role between reconstructing text and images. In vision, the decoder reconstructs \emph{pixels}, hence its output is of a lower \mbox{semantic} level than common recognition tasks. This is in contrast to language, where the decoder predicts missing \emph{words} that contain rich semantic information. While in BERT the decoder can be trivial (an MLP) \cite{Devlin2019}, we found that for images, the decoder design plays a key role in determining the semantic level of the learned latent representations.

Driven by this analysis, we present a simple, effective, and scalable form of a masked autoencoder (MAE) for visual representation learning. Our MAE masks random patches from the input image and reconstructs the missing patches in the pixel space. It has an \textit{asymmetric} encoder-decoder design. Our encoder operates only on the visible subset of patches (without mask tokens), and our decoder is lightweight and reconstructs the input from the latent representation along with mask tokens (Figure~\ref{fig:arch}). Shifting the mask tokens to the small decoder in our asymmetric encoder-decoder results in a large reduction in computation. Under this design, a very high masking ratio (\eg, 75\%) can achieve a win-win scenario: it optimizes accuracy while allowing the encoder to process only a small portion (\eg, 25\%) of patches. This can reduce overall pre-training time by 3$\x$ or more and likewise reduce memory consumption, enabling us to easily scale our MAE to large models.

Our MAE learns very high-capacity models that generalize well. With MAE pre-training, we can train data-hungry models like ViT-Large/-Huge \cite{Dosovitskiy2021} on ImageNet-1K with improved generalization performance. With a vanilla \mbox{ViT-Huge} model, we achieve 87.8\% accuracy when fine-tuned on ImageNet-1K. This outperforms all previous results that use only ImageNet-1K data. We also evaluate transfer learning on object detection, instance segmentation, and semantic segmentation. In these tasks, our pre-training achieves better results than its supervised pre-training counterparts, and more importantly, we observe significant gains by scaling up models. These observations are aligned with those witnessed in self-supervised pre-training in NLP \cite{Devlin2019, Radford2018, Radford2019, Brown2020} and we hope that they will enable our field to explore a similar trajectory.

\section{Related Work}\label{sec:related}

\paragraph{Masked language modeling} and its autoregressive counterparts, \eg, BERT \cite{Devlin2019} and GPT \cite{Radford2018, Radford2019, Brown2020}, are highly successful methods for pre-training in NLP. These methods hold out a portion of the input sequence and train models to predict the missing content. These methods have been shown to scale excellently \cite{Brown2020} and a large abundance of evidence indicates that these pre-trained representations generalize well to various downstream tasks.

\paragraph{Autoencoding} is a classical method for learning representations. It has an encoder that maps an input to a latent representation and a decoder that reconstructs the input. For example, PCA and k-means are autoencoders \cite{Hinton1994}. Denoising autoencoders (DAE) \cite{Vincent2008} are a class of autoencoders that corrupt an input signal and learn to reconstruct the original, uncorrupted signal. A series of methods can be thought of as a generalized DAE under different corruptions, \eg, masking pixels \cite{Vincent2010, Pathak2016, Chen2020c} or removing color channels \cite{Zhang2016}. Our MAE is a form of denoising autoencoding, but different from the classical DAE in numerous ways.

\paragraph{Masked image encoding} methods learn representations from images corrupted by masking. The pioneering work of \cite{Vincent2010} presents masking as a noise type in DAE. Context Encoder \cite{Pathak2016} inpaints large missing regions using convolutional networks. Motivated by the success in NLP, related recent methods \cite{Chen2020c, Dosovitskiy2021, Bao2021} are based on Transformers \cite{Vaswani2017}. iGPT \cite{Chen2020c} operates on sequences of pixels and predicts unknown pixels. The ViT paper \cite{Dosovitskiy2021} studies masked patch prediction for self-supervised learning. Most recently, BEiT \cite{Bao2021} proposes to predict discrete tokens \cite{Oord2017, Ramesh2021}.

\paragraph{Self-supervised learning} approaches have seen significant interest in computer vision, often focusing on different pretext tasks for pre-training \cite{Doersch2015, Wang2015a, Noroozi2016, Zhang2016, Pathak2017, Gidaris2018}. Recently, contrastive learning \cite{Becker1992, Hadsell2006} has been popular, \eg, \cite{Wu2018a, Oord2018, He2020, Chen2020}, which models image similarity and dissimilarity (or only similarity \cite{Grill2020, Chen2021}) between two or more views. Contrastive and related methods strongly depend on data augmentation \cite{Chen2020, Grill2020, Chen2021}. Autoencoding pursues a conceptually different direction, and it exhibits different behaviors as we will present.

\vspace{1mm}\section{Approach}\vspace{0.5mm}
\label{sec:approach}

Our masked autoencoder (MAE) is a simple autoencoding approach that reconstructs the original signal given its partial observation. Like all autoencoders, our approach has an encoder that maps the observed signal to a latent representation, and a decoder that reconstructs the original signal from the latent representation. Unlike classical autoencoders, we adopt an \emph{asymmetric} design that allows the encoder to operate only on the partial, observed signal (without mask tokens) and a lightweight decoder that reconstructs the full signal from the latent representation and mask tokens. Figure~\ref{fig:arch} illustrates the idea, introduced next.

\paragraph{Masking.} Following ViT \cite{Dosovitskiy2021}, we divide an image into regular non-overlapping patches. Then we sample a subset of patches and mask (\ie, remove) the remaining ones. Our sampling strategy is straightforward: we sample random patches without replacement, following a uniform distribution. We simply refer to this as ``random sampling".

Random sampling with a \textit{high} masking ratio (\ie, the ratio of removed patches) largely eliminates redundancy, thus creating a task that cannot be easily solved by extrapolation from visible neighboring patches (see Figures~\ref{fig:samples} -- \ref{fig:mask_generalization}). The uniform distribution prevents a potential center bias (\ie, more masked patches near the image center). Finally, the highly sparse input creates an opportunity for designing an efficient encoder, introduced next.

\paragraph{MAE encoder.} Our encoder is a ViT \cite{Dosovitskiy2021} but applied only on \emph{visible, unmasked patches}. Just as in a standard ViT, our encoder embeds patches by a linear projection with added positional embeddings, and then processes the resulting set via a series of Transformer blocks. However, our encoder only operates on a small subset (\eg, 25\%) of the full set. Masked patches are removed; no mask tokens are used. This allows us to train very large encoders with only a fraction of compute and memory. The full set is handled by a lightweight decoder, described next.

\paragraph{MAE decoder.} The input to the MAE decoder is the full set of tokens consisting of (i) encoded visible patches, and (ii) mask tokens. See Figure~\ref{fig:arch}. Each mask token \cite{Devlin2019} is a shared, learned vector that indicates the presence of a missing patch to be predicted. We add positional embeddings to all tokens in this full set; without this, mask tokens would have no information about their location in the image. The decoder has another series of Transformer blocks.

The MAE decoder is only used during pre-training to perform the image reconstruction task (only the encoder is used to produce image representations for recognition). Therefore, the decoder architecture can be flexibly designed in a manner that is \emph{independent} of the encoder design. We experiment with very small decoders, narrower and shallower than the encoder. For example, our default decoder has $<$10\% computation per token \vs the encoder. With this asymmetrical design, the full set of tokens are only processed by the lightweight decoder, which significantly reduces pre-training time.

\paragraph{Reconstruction target.} Our MAE reconstructs the input by predicting the \textit{pixel} values for each masked patch. Each element in the decoder's output is a vector of pixel values representing a patch. The last layer of the decoder is a linear projection whose number of output channels equals the number of pixel values in a patch. The decoder's output is reshaped to form a reconstructed image. Our loss function computes the mean squared error (MSE) between the reconstructed and original images in the pixel space. We compute the loss only on \mbox{masked} patches, similar to BERT \cite{Devlin2019}.\footnotemark

\footnotetext{Computing the loss only on masked patches differs from traditional denoising autoencoders \cite{Vincent2008} that compute the loss on all pixels. This choice is purely result-driven: computing the loss on all pixels leads to a slight decrease in accuracy (\eg, \app0.5\%).}

We also study a variant whose reconstruction target is the normalized pixel values of each masked patch. Specifically, we compute the mean and standard deviation of all pixels in a patch and use them to normalize this patch. Using normalized pixels as the reconstruction target improves representation quality in our experiments.

\paragraph{Simple implementation.} Our MAE pre-training can be implemented efficiently, and importantly, does not require any specialized sparse operations. First we generate a token for every input patch (by linear projection with an added positional embedding). Next we \emph{randomly shuffle} the list of tokens and \emph{remove} the last portion of the list, based on the masking ratio. This process produces a small subset of tokens for the encoder and is equivalent to sampling patches without replacement. After encoding, we append a list of mask tokens to the list of encoded patches, and \emph{unshuffle} this full list (inverting the random shuffle operation) to align all tokens with their targets. The decoder is applied to this full list (with positional embeddings added). As noted, no sparse operations are needed. This simple implementation introduces negligible overhead as the shuffling and unshuffling operations are fast.

\section{ImageNet Experiments}
\label{sec:exp}

We do self-supervised pre-training on the ImageNet-1K (IN1K) \cite{Deng2009} training set. Then we do supervised training to evaluate the representations with (i) end-to-end fine-tuning or (ii) linear probing. We report top-1 validation accuracy of a single 224$\times$224 crop. Details are in Appendix~\ref{app:impl_mae}.

\paragraph{Baseline: ViT-Large.} We use {ViT-Large} (ViT-L/16) \cite{Dosovitskiy2021} as the backbone in our ablation study. ViT-L is very big (an order of magnitude bigger than ResNet-50 \cite{He2016}) and tends to overfit. The following is a comparison between ViT-L trained from scratch \vs fine-tuned from our baseline MAE:
\begin{center}\vspace{-.2em}
\tablestyle{4pt}{1.05}
\begin{tabular}{x{68}x{60}x{60}}
scratch, original \cite{Dosovitskiy2021} & scratch, our impl. & baseline MAE \\
\shline
76.5 & 82.5 & 84.9
\end{tabular}\vspace{-.2em}
\end{center}
We note that it is nontrivial to train \textit{supervised} ViT-L from scratch and a good recipe with strong regularization is needed (82.5\%, see Appendix \ref{app:supervised_vit_large}). Even so, our MAE pre-training contributes a big improvement. Here fine-tuning is only for 50 epochs (\vs 200 from scratch), implying that the fine-tuning accuracy heavily depends on pre-training.

\begin{figure}[t]\centering
\vspace{-1em}
\includegraphics[width=.9\linewidth]{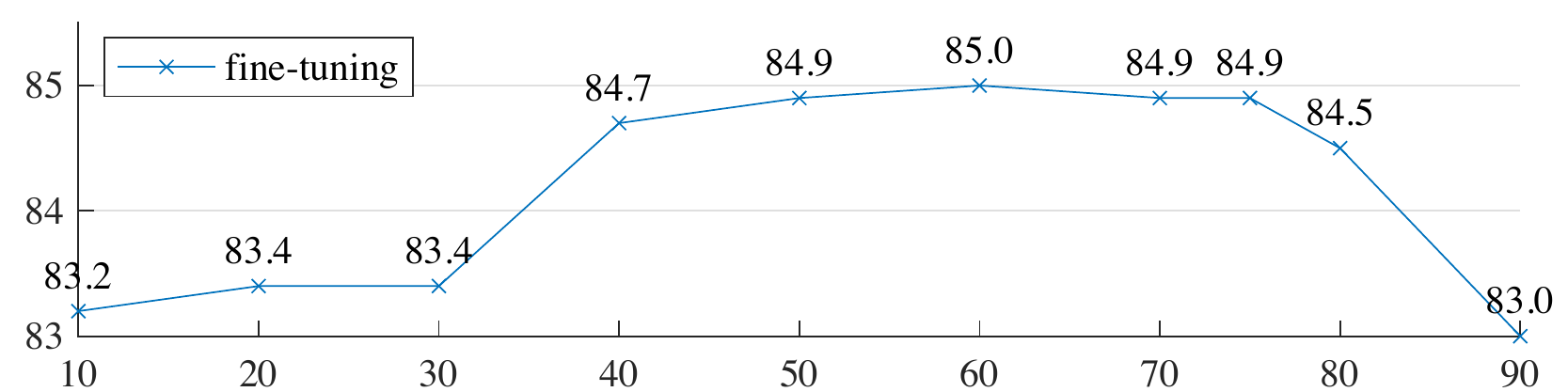}\\
\scriptsize masking ratio (\%) \\
\includegraphics[width=.9\linewidth]{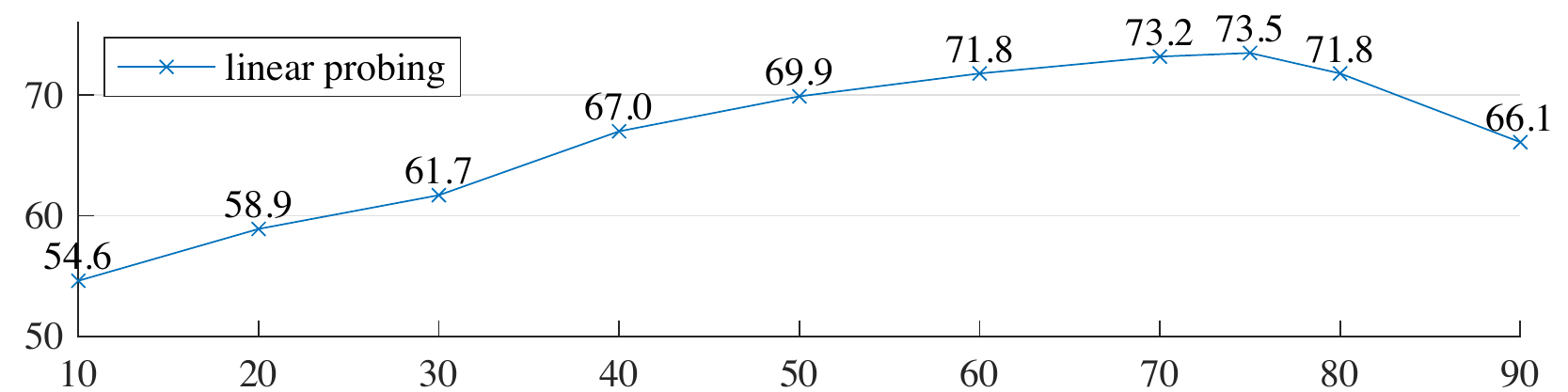}\\
\scriptsize masking ratio (\%) \\
\vspace{-.7em}
\caption{\textbf{Masking ratio}. A high masking ratio (75\%) works well for both fine-tuning (top) and linear probing (bottom). The y-axes are ImageNet-1K validation accuracy (\%) in all plots in this paper.
}
\label{fig:mask_ratio}
\vspace{-1em}
\end{figure}

\begin{table*}[t]
\vspace{-.2em}
\centering
\subfloat[
\textbf{Decoder depth}. A deep decoder can improve linear probing accuracy.
\label{tab:decoder_depth}
]{
\centering
\begin{minipage}{0.29\linewidth}{\begin{center}
\tablestyle{4pt}{1.05}
\begin{tabular}{x{18}x{24}x{24}}
blocks & ft & lin \\
\shline
1 & 84.8 & 65.5 \\
2 & \textbf{84.9} & 70.0 \\
4 & \textbf{84.9} & 71.9 \\
8 & \baseline{\textbf{84.9}} & \baseline{\textbf{73.5}} \\
12 & 84.4 & 73.3 \\
\end{tabular}
\end{center}}\end{minipage}
}
\hspace{2em}
\subfloat[
\textbf{Decoder width}. The decoder can be narrower than the encoder (1024-d).
\label{tab:decoder_width}
]{
\begin{minipage}{0.29\linewidth}{\begin{center}
\tablestyle{4pt}{1.05}
\begin{tabular}{x{18}x{24}x{24}}
dim & ft & lin \\
\shline
128 & \textbf{84.9} & 69.1 \\
256 & 84.8 & 71.3 \\
512 & \baseline{\textbf{84.9}} & \baseline{\textbf{73.5}} \\
768 & 84.4 & 73.1 \\
1024 & 84.3 & 73.1 \\
\end{tabular}
\end{center}}\end{minipage}
}
\hspace{2em}
\subfloat[
\textbf{Mask token}.
An encoder without mask tokens is more accurate and faster (Table~\ref{tab:wallclock}).
\label{tab:mask_token}
]{
\begin{minipage}{0.29\linewidth}{\begin{center}
\tablestyle{1pt}{1.05}
\begin{tabular}{y{56}x{24}x{24}z{24}}
case & ft & lin & FLOPs \\
\shline
{encoder w/ \texttt{[M]}} & 84.2 & 59.6 & 3.3$\times$ \\
{encoder {w/o} \texttt{[M]}} & \baseline{\textbf{84.9}} & \baseline{\textbf{73.5}} & \baseline{\textbf{1$\times$}} \\
\multicolumn{4}{c}{~}\\
\multicolumn{4}{c}{~}\\
\multicolumn{4}{c}{~}\\
\end{tabular}
\end{center}}\end{minipage}
}
\\
\centering
\vspace{.3em}
\subfloat[
\textbf{Reconstruction target}. Pixels as reconstruction targets
\label{tab:mae_target} are effective.
]{
\begin{minipage}{0.29\linewidth}{\begin{center}
\tablestyle{6pt}{1.05}
\begin{tabular}{y{54}x{24}x{24}}
case & ft & lin \\
\shline
pixel (w/o norm) & \baseline{84.9} & \baseline{73.5} \\
pixel (w/ norm) & \textbf{85.4} & \textbf{73.9} \\
PCA & 84.6 & 72.3 \\
dVAE token & 85.3 & 71.6 \\
\end{tabular}
\end{center}}\end{minipage}
}
\hspace{2em}
\subfloat[
\textbf{Data augmentation}. Our MAE works with minimal or no augmentation.
\label{tab:aug}
]{
\centering
\begin{minipage}{0.29\linewidth}{\begin{center}
\tablestyle{4pt}{1.05}
\begin{tabular}{y{54}x{22}x{22}}
case & ft & lin \\
\shline
none & 84.0 & 65.7 \\
crop, fixed size & 84.7 & 73.1 \\
crop, rand size & \baseline{\textbf{84.9}} & \baseline{\textbf{73.5}} \\
crop + color jit & 84.3 & 71.9 \\
\end{tabular}
\end{center}}\end{minipage}
}
\hspace{2em}
\subfloat[
\textbf{Mask sampling}. Random sampling works the best. See Figure~\ref{fig:mask_sampling} for visualizations.
\label{tab:mask_types}
]{
\begin{minipage}{0.29\linewidth}{\begin{center}
\tablestyle{1pt}{1.05}
\begin{tabular}{y{28}x{24}x{24}x{24}}
case & ratio & ft & lin \\
\shline
random & 75 & \baseline{\textbf{84.9}} & \baseline{\textbf{73.5}} \\
block & 50 & 83.9 & 72.3 \\
block & 75 & 82.8 & 63.9 \\
grid & 75 & 84.0 & 66.0 \\
\end{tabular}
\end{center}}\end{minipage}
}
\vspace{-.1em}
\caption{\textbf{MAE ablation experiments} with ViT-L/16 on ImageNet-1K. We report fine-tuning (ft) and linear probing (lin) accuracy (\%). If not specified, the default is: the decoder has depth 8 and width 512, the reconstruction target is unnormalized pixels, the data augmentation is random resized cropping, the masking ratio is 75\%, and the pre-training length is 800 epochs. Default settings are marked in \colorbox{baselinecolor}{gray}.}
\label{tab:ablations} \vspace{-.5em}
\end{table*}

\subsection{Main Properties}

We ablate our MAE using the default settings in Table~\ref{tab:ablations} (see caption). Several intriguing properties are observed.

\paragraph{Masking ratio.} Figure~\ref{fig:mask_ratio} shows the influence of the masking ratio. The optimal ratios are surprisingly high. The ratio of 75\% is good for both linear probing and fine-tuning. This behavior is in contrast with BERT \cite{Devlin2019}, whose typical masking ratio is 15\%. Our masking ratios are also much higher than those in related works \cite{Chen2020c,Dosovitskiy2021,Bao2021} in computer vision (20\% to 50\%).

The model \textit{infers} missing patches to produce different, yet plausible, outputs (Figure~\ref{fig:mask_generalization}). It makes sense of the gestalt of objects and scenes, which cannot be simply completed by extending lines or textures. We hypothesize that this reasoning-like behavior is linked to the learning of useful representations.

Figure~\ref{fig:mask_ratio} also shows that linear probing and fine-tuning results follow \textit{different} trends. For linear probing, the accuracy increases steadily with the masking ratio until the sweet point: the accuracy gap is up to $\app$20\% (54.6\% \vs 73.5\%). For fine-tuning, the results are less sensitive to the ratios, and a wide range of masking ratios (40--80\%) work well. All fine-tuning results in Figure~\ref{fig:mask_ratio} are better than training from scratch (82.5\%).

\paragraph{Decoder design.} Our MAE decoder can be flexibly designed, as studied in Table~\ref{tab:decoder_depth} and~\ref{tab:decoder_width}.

Table~\ref{tab:decoder_depth} varies the decoder depth (number of Transformer blocks). A sufficiently deep decoder is important for linear probing. This can be explained by the gap between a pixel reconstruction task and a recognition task: the last several layers in an autoencoder are more specialized for reconstruction, but are less relevant for recognition. A reasonably deep decoder can account for the reconstruction specialization, leaving the latent representations at a more abstract level. This design can yield up to 8\% improvement in linear probing (Table~\ref{tab:decoder_depth}, `lin'). However, if fine-tuning is used, the last layers of the encoder can be tuned to adapt to the recognition task. The decoder depth is less influential for improving fine-tuning (Table~\ref{tab:decoder_depth}, `ft').

Interestingly, our MAE with a \textit{single}-block decoder can perform strongly with fine-tuning (84.8\%). Note that a single Transformer block is the minimal requirement to propagate information from visible tokens to mask tokens. Such a small decoder can further speed up training.

In Table~\ref{tab:decoder_width} we study the decoder width (number of channels). We use 512-d by default, which performs well under fine-tuning and linear probing. A narrower decoder also works well with fine-tuning.

Overall, our default MAE decoder is lightweight. It has 8 blocks and a width of 512-d (\colorbox{baselinecolor}{gray} in Table~\ref{tab:ablations}). It only has 9\% FLOPs per token \vs ViT-L (24 blocks, 1024-d). As such, while the decoder processes all tokens, it is still a small fraction of the overall compute.

\begin{table}
\tablestyle{2pt}{1.1}
\begin{tabular}{y{56}x{36}x{32}x{28}x{28}}
encoder & dec. depth & ft acc & hours & speedup \\
\shline
\gc{ViT-L, w/ \texttt{[M]}} & \gc{8} & \gc{84.2} & \gc{42.4} & \gc{-} \\
ViT-L & 8 & 84.9 & 15.4 & 2.8$\times$ \\
ViT-L & 1 & 84.8 & 11.6 & \textbf{3.7}$\times$ \\
\hline
\gc{ViT-H, w/ \texttt{[M]}} & \gc{8} & \gc{-} & \gc{119.6$^\dagger$} & \gc{-} \\
ViT-H & 8 & 85.8 & 34.5 & 3.5$\times$ \\
ViT-H & 1 & 85.9 & 29.3 & \textbf{4.1}$\times$ \\
\end{tabular}
\vspace{-.7em}
\caption{\textbf{Wall-clock time} of our MAE training (800 epochs), benchmarked in 128 TPU-v3 cores with TensorFlow. The speedup is relative to the entry whose encoder has mask tokens (\gc{gray}). The decoder width is 512, and the mask ratio is 75\%. $^\dagger$: This entry is estimated by training ten epochs.}
\label{tab:wallclock}
\vspace{-.5em}
\end{table}

\paragraph{Mask token.} An important design of our MAE is to skip the mask token \texttt{[M]} in the encoder and apply it later in the lightweight decoder. Table~\ref{tab:mask_token} studies this design.

If the encoder \textit{uses} mask tokens, it performs \textit{worse}: its accuracy drops by 14\% in linear probing. In this case, there is a gap between pre-training and deploying: this encoder has a large portion of mask tokens in its input in pre-training, which does not exist in uncorrupted images. This gap may degrade accuracy in deployment. By removing the mask token from the encoder, we constrain the encoder to always see \textit{real} patches and thus improve accuracy.

Moreover, by skipping the mask token in the encoder, we greatly reduce training computation. In Table~\ref{tab:mask_token}, we reduce the overall training FLOPs by 3.3$\times$. This leads to a 2.8$\times$ wall-clock speedup in our implementation (see Table~\ref{tab:wallclock}). The wall-clock speedup is even bigger (3.5--4.1$\times$), for a smaller decoder (1-block), a larger encoder (\mbox{ViT-H}), or both. Note that the speedup can be $>$4$\times$ for a masking ratio of 75\%, partially because the self-attention complexity is quadratic. In addition, memory is greatly reduced, which can enable training even larger models or speeding up more by large-batch training. The time and memory efficiency makes our MAE favorable for training very large models.

\paragraph{Reconstruction target.} We compare different reconstruction targets in Table~\ref{tab:mae_target}. Our results thus far are based on pixels without (per-patch) normalization. Using pixels \textit{with} normalization improves accuracy. This per-patch normalization enhances the contrast locally. In another variant, we perform PCA in the patch space and use the largest PCA coefficients (96 here) as the target. Doing so degrades accuracy. Both experiments suggest that the high-frequency components are useful in our method.

We also compare an MAE variant that predicts \textit{tokens}, the target used in BEiT \cite{Bao2021}. Specifically for this variant, we use the DALLE pre-trained dVAE \cite{Ramesh2021} as the tokenizer, following \cite{Bao2021}. Here the MAE decoder predicts the token indices using cross-entropy loss. This tokenization improves fine-tuning accuracy by 0.4\% \vs unnormalized pixels, but has no advantage \vs normalized pixels. It also reduces linear probing accuracy. In \mbox{\S\ref{sec:transfer}} we further show that tokenization is not necessary in transfer learning.

Our \textit{pixel}-based MAE is much simpler than tokenization. The dVAE tokenizer requires one more pre-training stage, which may depend on extra data (250M images \cite{Ramesh2021}). The dVAE encoder is a large convolutional network (40\% FLOPs of ViT-L) and adds nontrivial overhead. Using pixels does not suffer from these problems.

\begin{figure}[t]
\centering
\includegraphics[width=0.99\linewidth]{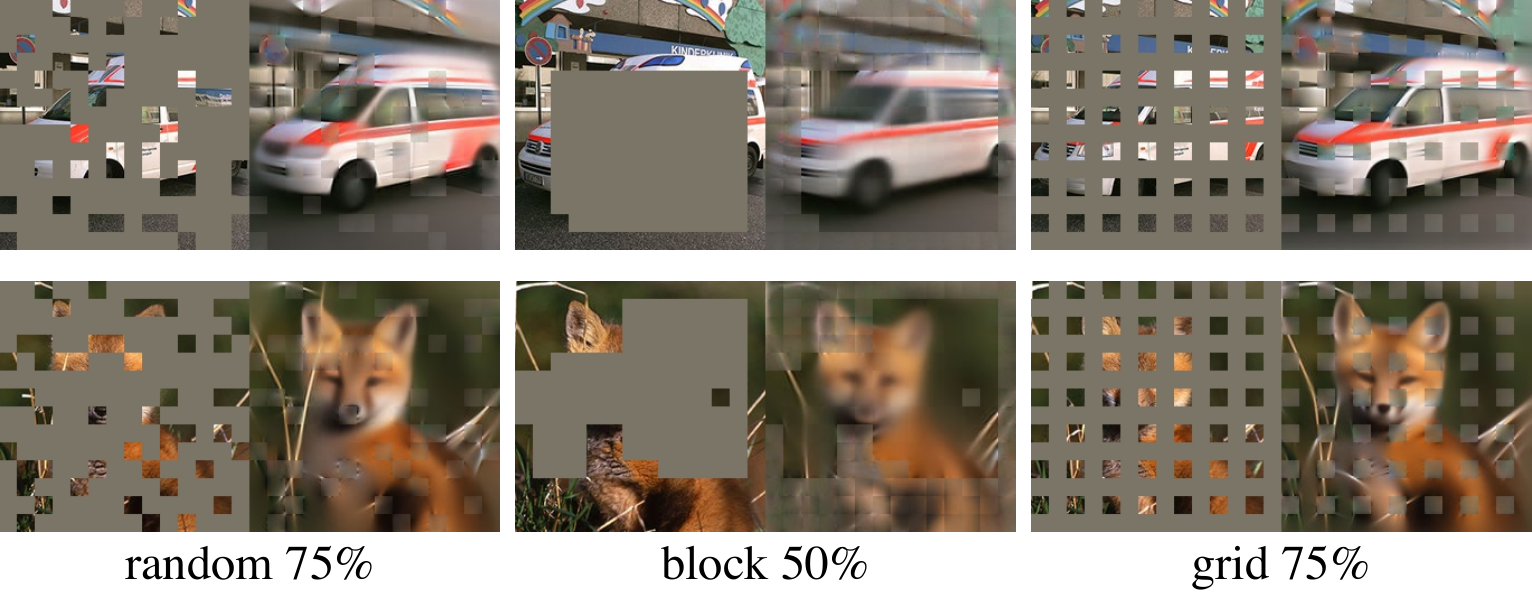}
\vspace{-.3em}
\caption{\textbf{Mask sampling strategies} determine the pretext task difficulty, influencing reconstruction quality and representations (Table~\ref{tab:mask_types}). Here each output is from an MAE trained with the specified masking strategy. {Left}: random sampling (our default). {Middle}: block-wise sampling \cite{Bao2021} that removes large random blocks. {Right}: grid-wise sampling that keeps one of every four patches. Images are from the validation set.}
\label{fig:mask_sampling}\vspace{.5em}
\end{figure}

\paragraph{Data augmentation.} Table~\ref{tab:aug} studies the influence of data augmentation on our MAE pre-training.

Our MAE works well using \textit{cropping-only} augmentation, either fixed-size or random-size (both having random horizontal flipping). Adding color jittering degrades the results and so we do not use it in other experiments.

Surprisingly, our MAE behaves decently even if using \textit{no data augmentation} (only center-crop, no flipping). This property is dramatically different from contrastive learning and related methods \cite{Wu2018a,He2020,Chen2020,Grill2020}, which heavily rely on data augmentation. It was observed \cite{Grill2020} that using cropping-only augmentation reduces the accuracy by 13\% and 28\% respectively for BYOL \cite{Grill2020} and SimCLR \cite{Chen2020}. In addition, there is no evidence that contrastive learning can work without augmentation: the two views of an image are the same and can easily satisfy a trivial solution.

In MAE, the role of data augmentation is mainly performed by random masking (ablated next). The masks are different for each iteration and so they generate new training samples regardless of data augmentation. The pretext task is made difficult by masking and requires less augmentation to regularize training.

\paragraph{Mask sampling strategy.} In Table~\ref{tab:mask_types} we compare different mask sampling strategies, illustrated in Figure~\ref{fig:mask_sampling}.

The \textit{block-wise} masking strategy, proposed in \cite{Bao2021}, tends to remove large blocks (Figure~\ref{fig:mask_sampling} middle). Our MAE with block-wise masking works reasonably well at a ratio of 50\%, but degrades at a ratio of 75\%. This task is harder than that of random sampling, as a higher training loss is observed. The reconstruction is also blurrier.

We also study \textit{grid-wise} sampling, which regularly keeps one of every four patches (Figure~\ref{fig:mask_sampling} right). This is an easier task and has lower training loss. The reconstruction is sharper. However, the representation quality is lower.

Simple random sampling works the best for our MAE. It allows for a higher masking ratio, which provides a greater speedup benefit while also enjoying good accuracy.

\begin{figure}[t]\centering
\includegraphics[width=0.99\linewidth]{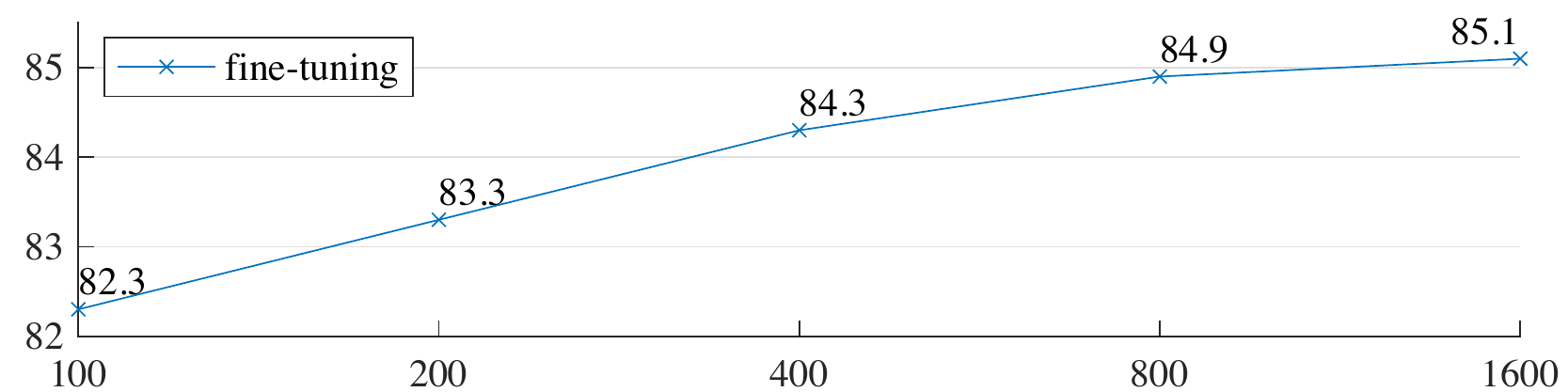}\\
\scriptsize epochs (log-scale) \\
\includegraphics[width=0.99\linewidth]{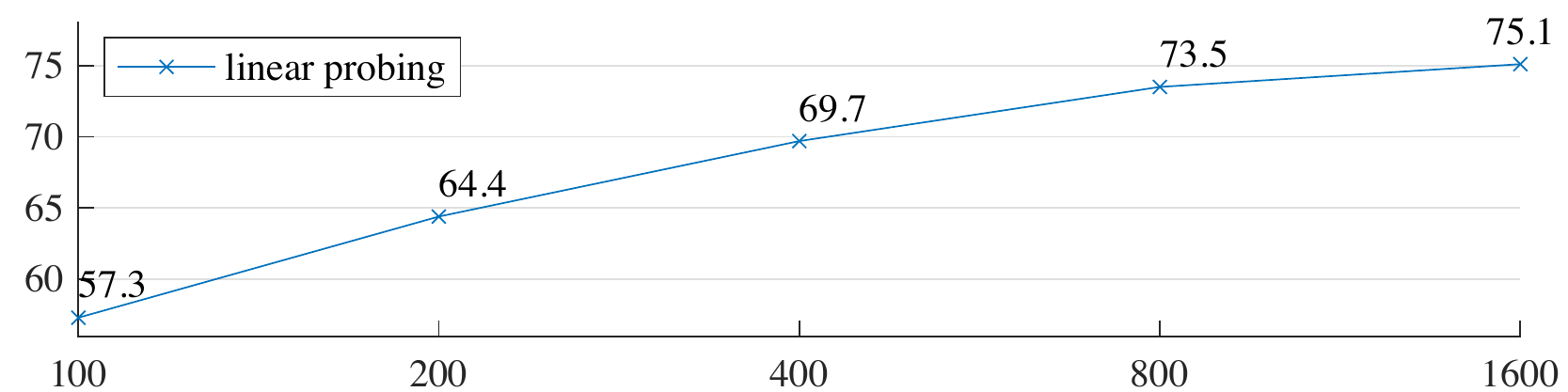}\\
\scriptsize epochs (log-scale) \\
\vspace{-.5em}
\caption{\textbf{Training schedules}. A longer training schedule gives a \mbox{noticeable} improvement. Here each point is a full training schedule. The model is ViT-L with the default setting in Table~\ref{tab:ablations}.}
\label{fig:schedule}
\vspace{-.5em}
\end{figure}

\paragraph{Training schedule.} Our ablations thus far are based on 800-epoch pre-training. Figure~\ref{fig:schedule} shows the influence of the training schedule length. The accuracy improves steadily with longer training. Indeed, we have not observed saturation of linear probing accuracy even at 1600 epochs. This behavior is unlike contrastive learning methods, \eg, MoCo~v3 \cite{Chen2021a} saturates at 300 epochs for ViT-L. Note that the MAE encoder only sees 25\% of patches per epoch, while in contrastive learning the encoder sees 200\% (two-crop) or even more (multi-crop) patches per epoch.

\begin{table}
\vspace{-.5em}
\tablestyle{3pt}{1.1}
\begin{tabular}{l l x{24}x{24}x{24}x{24}}
\multirow{1}{*}{method} &
\multirow{1}{*}{pre-train data}
& \multicolumn{1}{c}{ViT-B} & \multicolumn{1}{c}{ViT-L} & \multicolumn{1}{c}{ViT-H} & \multicolumn{1}{c}{ViT-H$_\text{448}$} \\
\shline
\gc{scratch, our impl.} & \gc{-} & \gc{82.3} & \gc{82.6} & \gc{83.1} & \gc{-} \\
DINO \cite{Caron2021} & \scriptsize IN1K & 82.8 & - & - & - \\
MoCo v3 \cite{Chen2021a} & \scriptsize IN1K & 83.2 & 84.1 & - & - \\
BEiT \cite{Bao2021} & \scriptsize IN1K+DALLE & 83.2 & 85.2 & - & - \\
\hline
MAE & \scriptsize IN1K & \underline{83.6} & \underline{85.9} & \underline{86.9} & \textbf{87.8} \\
\end{tabular}
\vspace{-.8em}
\caption{\textbf{Comparisons with previous results on ImageNet-1K}. The pre-training data is the ImageNet-1K training set (except the tokenizer in BEiT was pre-trained on 250M DALLE data \cite{Ramesh2021}). All self-supervised methods are evaluated by end-to-end fine-tuning. The ViT models are B/16, L/16, H/14 \cite{Dosovitskiy2021}. The best for each column is underlined. All results are on an image size of 224, except for ViT-H with an extra result on 448. Here our MAE reconstructs normalized pixels and is pre-trained for 1600 epochs.}
\label{tab:imagenet_e2e}
\end{table}

\subsection{Comparisons with Previous Results}

\paragraph{Comparisons with self-supervised methods.} In Table~\ref{tab:imagenet_e2e} we compare the fine-tuning results of self-supervised ViT models. For ViT-B, all methods perform closely. For \mbox{ViT-L}, the gaps among methods are bigger, suggesting that a challenge for bigger models is to reduce overfitting.

\begin{figure}[t]\centering
\vspace{-.7em}
\begin{overpic}[percent,width=.98\linewidth]{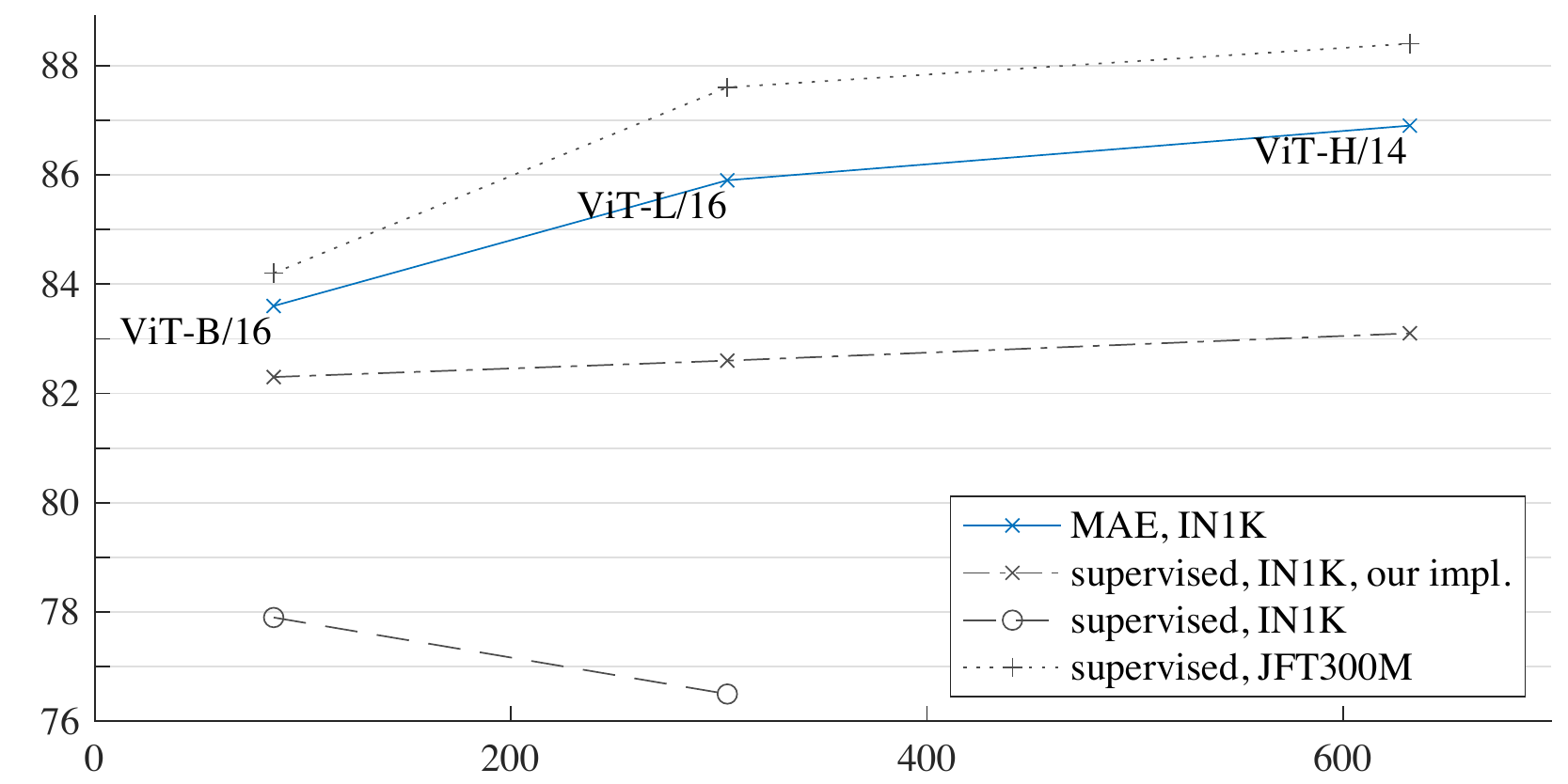}
\put(86.8,9.8){\tiny \cite{Dosovitskiy2021}} 
\put(90.8,6.8){\tiny \cite{Dosovitskiy2021}} 
\end{overpic}
\vspace{-.5em}
\scriptsize params (M) \\
\vspace{-.5em}
\caption{\textbf{MAE pre-training \vs supervised pre-training}, evaluated by fine-tuning in ImageNet-1K (224 size). We compare with the original ViT results \cite{Dosovitskiy2021} trained in IN1K or JFT300M.}
\label{fig:model_size} \vspace{-.7em}
\end{figure}

Our MAE can scale up easily and has shown steady improvement from bigger models. We obtain 86.9\% accuracy using \mbox{ViT-H} (224 size). By fine-tuning with a 448 size, we achieve \textbf{87.8}\% accuracy, \textit{using only IN1K data}. The previous best accuracy, among all methods using only IN1K data, is 87.1\% (512 size) \cite{Yuan2021}, based on advanced networks. We improve over the state-of-the-art by a nontrivial margin in the highly competitive benchmark of IN1K (no external data). Our result is based on \textit{vanilla} ViT, and we expect advanced networks will perform better.

Comparing with BEiT \cite{Bao2021}, our MAE is \textit{more accurate} while being \textit{simpler} and \textit{faster}. Our method reconstructs pixels, in contrast to BEiT that predicts tokens: BEiT reported a 1.8\% degradation \cite{Bao2021} when reconstructing pixels with \mbox{ViT-B}.\footnotemark~We do not need dVAE pre-training. Moreover, our MAE is considerably faster (3.5$\times$ per epoch) than BEiT, for the reason as studied in Table~\ref{tab:mask_token}.

\footnotetext{We observed the degradation also in BEiT with ViT-L: it produces 85.2\% (tokens) and 83.5\% (pixels), reproduced from the official code.}

The MAE models in Table~\ref{tab:imagenet_e2e} are pre-trained for 1600 epochs for better accuracy (Figure~\ref{fig:schedule}). Even so, our total pre-training time is \textit{less} than the other methods when trained on the same hardware. For example, training \mbox{ViT-L} on 128 TPU-v3 cores, our MAE's training time is 31 hours for 1600 epochs and MoCo v3's is 36 hours for 300 epochs \cite{Chen2021a}.

\paragraph{Comparisons with supervised pre-training.} In the original ViT paper \cite{Dosovitskiy2021}, ViT-L degrades when trained in IN1K. Our implementation of supervised training (see \ref{app:supervised_vit_large}) works better, but accuracy saturates. See Figure~\ref{fig:model_size}. 

Our MAE pre-training, using only IN1K, can generalize better: the gain over training from scratch is bigger for higher-capacity models. It follows a trend similar to the \mbox{JFT-300M} \textit{supervised} pre-training in \cite{Dosovitskiy2021}. This comparison shows that our MAE can help scale up model sizes.

\begin{figure}[t]\centering
\vspace{-.7em}
\includegraphics[width=.9\linewidth]{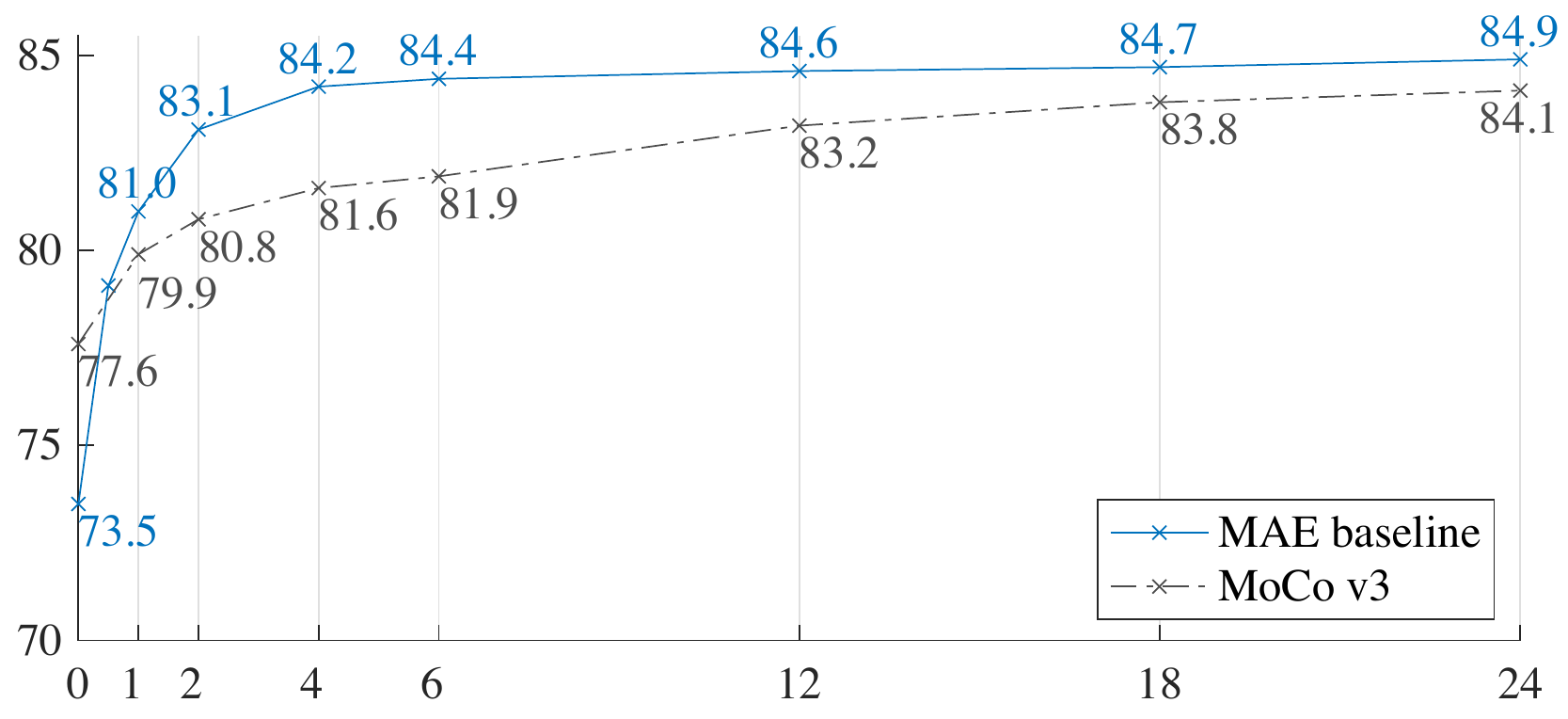} \\
\vspace{-.5em}
{\scriptsize \# blocks fine-tuned} \\
\vspace{-.3em}
\caption{\textbf{Partial fine-tuning} results of ViT-L \wrt the number of fine-tuned Transformer blocks under the default settings from Table~\ref{tab:ablations}. Tuning 0 blocks is linear probing; 24 is full fine-tuning. Our MAE representations are less linearly separable, but are consistently better than MoCo v3 if one or more blocks are tuned.}
\label{fig:partial_ft} \vspace{-1em}
\end{figure}

\subsection{Partial Fine-tuning}
\label{sec:partial_ft}

Table~\ref{tab:ablations} shows that linear probing and fine-tuning results are largely \textit{uncorrelated}. Linear probing has been a popular protocol in the past few years; however, it misses the opportunity of pursuing \textit{strong but non-linear} features---which is indeed a strength of deep learning. As a middle ground, we study a \textit{partial fine-tuning} protocol: fine-tune the last several layers while freezing the others. This protocol was also used in early works, \eg, \cite{Yosinski2014,Zhang2016,Noroozi2016}.

Figure~\ref{fig:partial_ft} shows the results. Notably, fine-tuning only \textit{one} Transformer block boosts the accuracy significantly from 73.5\% to 81.0\%. Moreover, if we fine-tune only ``half" of the last block (\ie, its MLP sub-block), we can get 79.1\%, much better than linear probing. This variant is essentially fine-tuning an MLP head. Fine-tuning a few blocks (\eg, 4 or 6) can achieve accuracy close to full fine-tuning.

In Figure~\ref{fig:partial_ft} we also compare with MoCo v3 \cite{Chen2021a}, a contrastive method with ViT-L results available. MoCo v3 has higher linear probing accuracy; however, all of its partial fine-tuning results are worse than MAE. The gap is 2.6\% when tuning 4 blocks. While the MAE representations are less linearly separable, they are stronger \textit{non-linear} features and perform well when a non-linear head is tuned.

These observations suggest that linear separability is not the sole metric for evaluating representation quality. It has also been observed (\eg, \cite{Chen2021}) that linear probing is not well \mbox{correlated} with transfer learning performance, \eg, for object detection. To our knowledge, linear evaluation is not often used in NLP for benchmarking pre-training.

\section{Transfer Learning Experiments}\label{sec:transfer}

\begin{table}[t]
\vspace{-1.5em} 
\tablestyle{5pt}{1.05}
\begin{tabular}{llcccc}
 & & \multicolumn{2}{c}{AP$^\text{box}$} & \multicolumn{2}{c}{\gc{AP$^\text{mask}$}} \\
method & pre-train data & ViT-B & ViT-L & \gc{ViT-B} & \gc{ViT-L} \\
\shline
supervised & \scriptsize IN1K w/ labels & 47.9 & 49.3 & \gc{42.9} & \gc{43.9} \\
MoCo v3 & \scriptsize IN1K & 47.9 & 49.3 & \gc{42.7} & \gc{44.0} \\
BEiT & \scriptsize IN1K+{DALLE} & 49.8 & \textbf{53.3} & \gc{44.4} & \gc{47.1} \\
\hline
MAE & \scriptsize IN1K & \textbf{50.3} & \textbf{53.3} & \gc{\textbf{44.9}} & \gc{\textbf{47.2}} \\
\end{tabular}
\vspace{-.7em}
\caption{\textbf{COCO object detection and segmentation} using a ViT Mask R-CNN baseline. All entries are based on our implementation. Self-supervised entries use IN1K data \textit{without} labels. Mask AP follows a similar trend as box AP.}
\label{tab:coco} \vspace{-1em}
\end{table}

We evaluate transfer learning in downstream tasks using the pre-trained models in Table~\ref{tab:imagenet_e2e}.

\paragraph{Object detection and segmentation.} We fine-tune Mask R-CNN \cite{He2017} end-to-end on COCO \cite{Lin2014}. The ViT backbone is adapted for use with FPN~\cite{Lin2017} (see \ref{app:coco}). We apply this approach for all entries in Table~\ref{tab:coco}. We report box AP for object detection and mask AP for instance segmentation.

Compared to supervised pre-training, our MAE performs better under all configurations (Table~\ref{tab:coco}). With the smaller ViT-B, our MAE is 2.4 points higher than \textit{supervised} pre-training (50.3 \vs 47.9, AP$^\text{box}$). More significantly, with the larger ViT-L, our MAE pre-training outperforms supervised pre-training by 4.0 points (53.3 \vs 49.3).

The \textit{pixel}-based MAE is better than or on par with the \textit{token}-based BEiT, while MAE is much simpler and faster. Both MAE and BEiT are better than MoCo v3 and MoCo v3 is on par with supervised pre-training.

\paragraph{Semantic segmentation.} We experiment on ADE20K \cite{Zhou2019} using UperNet \cite{Xiao2018} (see \ref{app:ade20k}). Table~\ref{tab:ade20k} shows that our pre-training significantly improves results over \textit{supervised} pre-training, \eg, by 3.7 points for ViT-L. Our pixel-based MAE also outperforms the token-based BEiT. These observations are consistent with those in COCO.

\paragraph{Classification tasks.} \mbox{Table~\ref{tab:cls_transfer}} studies transfer learning on the iNaturalists \cite{VanHorn2018} and Places \cite{Zhou2014} tasks (see \ref{app:class}). On iNat, our method shows strong scaling behavior: accuracy improves considerably with bigger models. Our results surpass the previous best results \textit{by large margins}. On Places, our MAE outperforms the previous best results \cite{Goyal2021,Mahajan2018}, which were obtained via pre-training on billions of images.

\paragraph{Pixels \vs tokens.} Table~\ref{tab:pixel_vs_token} compares pixels \vs tokens as the MAE reconstruction target. While using dVAE tokens is better than using \textit{unnormalized} pixels, it is statistically similar to using \textit{normalized} pixels across all cases we tested. It again shows that tokenization is not necessary for our MAE.

\begin{table}[t]
\vspace{-.5em}
\tablestyle{8pt}{1.05}
\begin{tabular}{llcc}
method & pre-train data & ViT-B & ViT-L \\
\shline
supervised & \scriptsize IN1K w/ labels & 47.4 & 49.9 \\
MoCo v3 & \scriptsize IN1K & 47.3 & 49.1 \\
BEiT & \scriptsize IN1K+{DALLE} & 47.1 & 53.3 \\
\hline
MAE & \scriptsize IN1K & \textbf{48.1} & \textbf{53.6} \\
\end{tabular}
\vspace{-.7em}
\caption{\textbf{ADE20K semantic segmentation} (mIoU) using UperNet. BEiT results are reproduced using the official code. Other entries are based on our implementation. Self-supervised entries use IN1K data \textit{without} labels.}
\label{tab:ade20k} \vspace{-.5em}
\end{table}

\begin{table}[t]
\tablestyle{5pt}{1.05}
\begin{tabular}{l x{24}x{24}x{24}x{24}y{36}}
\multirow{1}{*}{dataset}
& \multicolumn{1}{c}{ViT-B} & \multicolumn{1}{c}{ViT-L} & \multicolumn{1}{c}{ViT-H} & \multicolumn{1}{c}{ViT-H$_\text{448}$} & \gc{prev best} \\
\shline
iNat 2017 & 70.5 & 75.7 & 79.3 & \textbf{83.4} & \gc{75.4} \cite{Touvron2019} \\
iNat 2018 & 75.4 & 80.1 & 83.0 & \textbf{86.8} & \gc{81.2} \cite{Touvron2021b} \\
iNat 2019 & 80.5 & 83.4 & 85.7 & \textbf{88.3} & \gc{84.1} \cite{Touvron2021b} \\
Places205 & 63.9 & 65.8 & 65.9 & \textbf{66.8} & \gc{66.0} \cite{Goyal2021}$^\dagger$ \\
Places365 & 57.9 & 59.4 & 59.8 & \textbf{60.3} & \gc{58.0} \cite{Mahajan2018}$^\ddagger$ \\
\end{tabular}
\vspace{-.8em}
\caption{\textbf{Transfer learning accuracy on classification datasets}, using MAE pre-trained on IN1K and then fine-tuned.
We provide system-level comparisons with the previous best results.
\\{\scriptsize $^\dagger$: pre-trained on 1 billion images. $^\ddagger$: pre-trained on 3.5 billion images.}
}
\label{tab:cls_transfer}
\vspace{-.5em}
\end{table}

\begin{table}[t]
\tablestyle{3.5pt}{1.05}
\begin{tabular}{l|rrr|rr|rr}
& \multicolumn{3}{c|}{\scriptsize IN1K}
& \multicolumn{2}{c|}{\scriptsize COCO}
& \multicolumn{2}{c}{\scriptsize ADE20K} \\
& \scriptsize ViT-B & \scriptsize ViT-L & \scriptsize ViT-H
& \scriptsize ViT-B & \scriptsize ViT-L
& \scriptsize ViT-B & \scriptsize ViT-L \\
\shline
pixel (w/o norm) & 83.3 & 85.1 & 86.2 & 49.5 & 52.8 & 48.0 & 51.8 \\
pixel (w/ norm) & 83.6 & 85.9 & 86.9 & 50.3 & 53.3 & 48.1 & 53.6 \\
\hline
dVAE token & 83.6 & 85.7 & 86.9 & 50.3 & 53.2 & 48.1 & 53.4 \\
$\triangle$ & 0.0 & -0.2 & 0.0 & 0.0 & -0.1 & 0.0 & -0.2
\end{tabular}
\vspace{-.7em}
\caption{\textbf{Pixels \vs tokens} as the MAE reconstruction target. $\triangle$ is the difference between using dVAE tokens and using normalized pixels. The difference is statistically insignificant.}
\label{tab:pixel_vs_token} \vspace{-0.5em}
\end{table}

\section{Discussion and Conclusion}

Simple algorithms that scale well are the core of deep learning. In NLP, simple self-supervised learning methods (\eg, \cite{Radford2018, Devlin2019, Radford2019, Brown2020}) enable benefits from exponentially scaling models. In computer vision, practical pre-training paradigms are dominantly supervised (\eg \cite{Krizhevsky2012,Simonyan2015,He2016,Dosovitskiy2021}) despite progress in self-supervised learning. In this study, we observe on ImageNet and in transfer learning that an autoencoder---a simple self-supervised method similar to techniques in NLP---provides scalable benefits. Self-supervised learning in vision may now be embarking on a similar trajectory as in NLP.

On the other hand, we note that images and languages are \textit{signals of a different nature} and this difference must be addressed carefully. Images are merely recorded light \mbox{\textit{without}} a semantic decomposition into the visual analogue of words. Instead of attempting to remove objects, we remove random patches that most likely do \textit{not} form a semantic segment. Likewise, our MAE reconstructs pixels, which are \emph{not} semantic entities. Nevertheless, we observe (\eg, Figure \ref{fig:mask_generalization}) that our MAE infers complex, holistic reconstructions, suggesting it has learned numerous visual concepts, \ie, semantics. We hypothesize that this behavior occurs by way of a rich hidden representation inside the MAE. We hope this perspective will inspire future work.

\paragraph{Broader impacts.} The proposed method predicts content based on learned statistics of the training dataset and as such will reflect biases in those data, including ones with negative societal impacts. The model may generate inexistent content. These issues warrant further research and consideration when building upon this work to generate images.

{
\fontsize{8.2pt}{9.84pt}\selectfont
\bibliographystyle{ieee_fullname}\bibliography{mae}}

\begin{thebibliography}{10}\itemsep=-1pt

\bibitem{Ba2016}
Jimmy~Lei Ba, Jamie~Ryan Kiros, and Geoffrey~E Hinton.
\newblock Layer normalization.
\newblock {\em arXiv:1607.06450}, 2016.

\bibitem{Bao2021}
Hangbo Bao, Li Dong, and Furu Wei.
\newblock {BEiT}: {BERT} pre-training of image transformers.
\newblock {\em arXiv:2106.08254}, 2021.
\newblock \emph{Accessed in June 2021}.

\bibitem{Becker1992}
Suzanna Becker and Geoffrey~E Hinton.
\newblock Self-organizing neural network that discovers surfaces in random-dot
  stereograms.
\newblock {\em Nature}, 1992.

\bibitem{Brown2020}
Tom Brown, Benjamin Mann, Nick Ryder, Melanie Subbiah, Jared~D Kaplan, Prafulla
  Dhariwal, Arvind Neelakantan, Pranav Shyam, Girish Sastry, Amanda Askell,
  Sandhini Agarwal, Ariel Herbert-Voss, Gretchen Krueger, Tom Henighan, Rewon
  Child, Aditya Ramesh, Daniel Ziegler, Jeffrey Wu, Clemens Winter, Chris
  Hesse, Mark Chen, Eric Sigler, Mateusz Litwin, Scott Gray, Benjamin Chess,
  Jack Clark, Christopher Berner, Sam McCandlish, Alec Radford, Ilya Sutskever,
  and Dario Amodei.
\newblock Language models are few-shot learners.
\newblock In {\em NeurIPS}, 2020.

\bibitem{Caron2021}
Mathilde Caron, Hugo Touvron, Ishan Misra, Herv{\'e} J{\'e}gou, Julien Mairal,
  Piotr Bojanowski, and Armand Joulin.
\newblock Emerging properties in self-supervised vision transformers.
\newblock In {\em ICCV}, 2021.

\bibitem{Chen2020c}
Mark Chen, Alec Radford, Rewon Child, Jeffrey Wu, Heewoo Jun, David Luan, and
  Ilya Sutskever.
\newblock Generative pretraining from pixels.
\newblock In {\em ICML}, 2020.

\bibitem{Chen2020}
Ting Chen, Simon Kornblith, Mohammad Norouzi, and Geoffrey Hinton.
\newblock A simple framework for contrastive learning of visual
  representations.
\newblock In {\em ICML}, 2020.

\bibitem{Chen2021}
Xinlei Chen and Kaiming He.
\newblock Exploring simple {Siamese} representation learning.
\newblock In {\em CVPR}, 2021.

\bibitem{Chen2021a}
Xinlei Chen, Saining Xie, and Kaiming He.
\newblock An empirical study of training self-supervised {Vision Transformers}.
\newblock In {\em ICCV}, 2021.

\bibitem{Clark2020}
Kevin Clark, Minh-Thang Luong, Quoc~V Le, and Christopher~D Manning.
\newblock {ELECTRA}: Pre-training text encoders as discriminators rather than
  generators.
\newblock In {\em ICLR}, 2020.

\bibitem{Cortes1995}
Corinna Cortes and Vladimir Vapnik.
\newblock Support-vector networks.
\newblock {\em Machine learning}, 1995.

\bibitem{Cubuk2020}
Ekin~D Cubuk, Barret Zoph, Jonathon Shlens, and Quoc~V Le.
\newblock Randaugment: Practical automated data augmentation with a reduced
  search space.
\newblock In {\em CVPR Workshops}, 2020.

\bibitem{Deng2009}
Jia Deng, Wei Dong, Richard Socher, Li-Jia Li, Kai Li, and Li Fei-Fei.
\newblock {ImageNet: A large-scale hierarchical image database}.
\newblock In {\em CVPR}, 2009.

\bibitem{Devlin2019}
Jacob Devlin, Ming-Wei Chang, Kenton Lee, and Kristina Toutanova.
\newblock {BERT}: Pre-training of deep bidirectional transformers for language
  understanding.
\newblock In {\em NAACL}, 2019.

\bibitem{Doersch2015}
Carl Doersch, Abhinav Gupta, and Alexei~A Efros.
\newblock Unsupervised visual representation learning by context prediction.
\newblock In {\em ICCV}, 2015.

\bibitem{Dosovitskiy2021}
Alexey Dosovitskiy, Lucas Beyer, Alexander Kolesnikov, Dirk Weissenborn,
  Xiaohua Zhai, Thomas Unterthiner, Mostafa Dehghani, Matthias Minderer, Georg
  Heigold, Sylvain Gelly, Jakob Uszkoreit, and Neil Houlsby.
\newblock An image is worth 16x16 words: Transformers for image recognition at
  scale.
\newblock In {\em ICLR}, 2021.

\bibitem{Gidaris2018}
Spyros Gidaris, Praveer Singh, and Nikos Komodakis.
\newblock Unsupervised representation learning by predicting image rotations.
\newblock In {\em ICLR}, 2018.

\bibitem{Glorot2010}
Xavier Glorot and Yoshua Bengio.
\newblock Understanding the difficulty of training deep feedforward neural
  networks.
\newblock In {\em AISTATS}, 2010.

\bibitem{Goyal2021}
Priya Goyal, Mathilde Caron, Benjamin Lefaudeux, Min Xu, Pengchao Wang, Vivek
  Pai, Mannat Singh, Vitaliy Liptchinsky, Ishan Misra, Armand Joulin, and Piotr
  Bojanowski.
\newblock Self-supervised pretraining of visual features in the wild.
\newblock {\em arXiv:2103.01988}, 2021.

\bibitem{Goyal2017}
Priya Goyal, Piotr Doll{\'a}r, Ross Girshick, Pieter Noordhuis, Lukasz
  Wesolowski, Aapo Kyrola, Andrew Tulloch, Yangqing Jia, and Kaiming He.
\newblock Accurate, large minibatch {SGD}: Training {ImageNet} in 1 hour.
\newblock {\em arXiv:1706.02677}, 2017.

\bibitem{Grill2020}
Jean-Bastien Grill, Florian Strub, Florent Altch\'{e}, Corentin Tallec, Pierre
  Richemond, Elena Buchatskaya, Carl Doersch, Bernardo Avila~Pires, Zhaohan
  Guo, Mohammad Gheshlaghi~Azar, Bilal Piot, Koray Kavukcuoglu, Remi Munos, and
  Michal Valko.
\newblock Bootstrap your own latent - a new approach to self-supervised
  learning.
\newblock In {\em NeurIPS}, 2020.

\bibitem{Hadsell2006}
Raia Hadsell, Sumit Chopra, and Yann LeCun.
\newblock Dimensionality reduction by learning an invariant mapping.
\newblock In {\em CVPR}, 2006.

\bibitem{He2020}
Kaiming He, Haoqi Fan, Yuxin Wu, Saining Xie, and Ross Girshick.
\newblock Momentum contrast for unsupervised visual representation learning.
\newblock In {\em CVPR}, 2020.

\bibitem{He2017}
Kaiming He, Georgia Gkioxari, Piotr Doll{\'a}r, and Ross Girshick.
\newblock {Mask R-CNN}.
\newblock In {\em ICCV}, 2017.

\bibitem{He2016}
Kaiming He, Xiangyu Zhang, Shaoqing Ren, and Jian Sun.
\newblock Deep residual learning for image recognition.
\newblock In {\em CVPR}, 2016.

\bibitem{Hendrycks2021a}
Dan Hendrycks, Steven Basart, Norman Mu, Saurav Kadavath, Frank Wang, Evan
  Dorundo, Rahul Desai, Tyler Zhu, Samyak Parajuli, Mike Guo, et~al.
\newblock The many faces of robustness: A critical analysis of
  out-of-distribution generalization.
\newblock In {\em ICCV}, 2021.

\bibitem{Hendrycks2019}
Dan Hendrycks and Thomas Dietterich.
\newblock Benchmarking neural network robustness to common corruptions and
  perturbations.
\newblock In {\em ICLR}, 2019.

\bibitem{Hendrycks2021}
Dan Hendrycks, Kevin Zhao, Steven Basart, Jacob Steinhardt, and Dawn Song.
\newblock Natural adversarial examples.
\newblock In {\em CVPR}, 2021.

\bibitem{Hinton1994}
Geoffrey~E Hinton and Richard~S Zemel.
\newblock Autoencoders, minimum description length, and helmholtz free energy.
\newblock In {\em NeurIPS}, 1994.

\bibitem{Huang2016}
Gao Huang, Yu Sun, Zhuang Liu, Daniel Sedra, and Kilian~Q Weinberger.
\newblock Deep networks with stochastic depth.
\newblock In {\em ECCV}, 2016.

\bibitem{Ioffe2015}
Sergey Ioffe and Christian Szegedy.
\newblock Batch normalization: Accelerating deep network training by reducing
  internal covariate shift.
\newblock In {\em ICML}, 2015.

\bibitem{Kim2021}
Insoo Kim, Seungju Han, Ji-won Baek, Seong-Jin Park, Jae-Joon Han, and Jinwoo
  Shin.
\newblock Quality-agnostic image recognition via invertible decoder.
\newblock In {\em CVPR}, 2021.

\bibitem{Krizhevsky2012}
Alex Krizhevsky, Ilya Sutskever, and Geoff Hinton.
\newblock Imagenet classification with deep convolutional neural networks.
\newblock In {\em NeurIPS}, 2012.

\bibitem{LeCun1989}
Yann LeCun, Bernhard Boser, John~S Denker, Donnie Henderson, Richard~E Howard,
  Wayne Hubbard, and Lawrence~D Jackel.
\newblock Backpropagation applied to handwritten zip code recognition.
\newblock {\em Neural computation}, 1989.

\bibitem{Li2021}
Yanghao Li, Saining Xie, Xinlei Chen, Piotr Doll{\'a}r, Kaiming He, and Ross
  Girshick.
\newblock Benchmarking detection transfer learning with vision transformers.
\newblock {\em In preparation}, 2021.

\bibitem{Lin2017}
Tsung-Yi Lin, Piotr Doll{\'a}r, Ross Girshick, Kaiming He, Bharath Hariharan,
  and Serge Belongie.
\newblock Feature pyramid networks for object detection.
\newblock In {\em CVPR}, 2017.

\bibitem{Lin2014}
Tsung-Yi Lin, Michael Maire, Serge Belongie, James Hays, Pietro Perona, Deva
  Ramanan, Piotr Doll{\'a}r, and C~Lawrence Zitnick.
\newblock {Microsoft COCO: Common objects in context}.
\newblock In {\em ECCV}, 2014.

\bibitem{Loshchilov2016}
Ilya Loshchilov and Frank Hutter.
\newblock {SGDR}: Stochastic gradient descent with warm restarts.
\newblock In {\em ICLR}, 2017.

\bibitem{Loshchilov2019}
Ilya Loshchilov and Frank Hutter.
\newblock Decoupled weight decay regularization.
\newblock In {\em ICLR}, 2019.

\bibitem{Mahajan2018}
Dhruv Mahajan, Ross Girshick, Vignesh Ramanathan, Kaiming He, Manohar Paluri,
  Yixuan Li, Ashwin Bharambe, and Laurens van~der Maaten.
\newblock Exploring the limits of weakly supervised pretraining.
\newblock In {\em ECCV}, 2018.

\bibitem{Mao2021}
Xiaofeng Mao, Gege Qi, Yuefeng Chen, Xiaodan Li, Ranjie Duan, Shaokai Ye, Yuan
  He, and Hui Xue.
\newblock Towards robust vision transformer.
\newblock {\em arXiv:2105.07926}, 2021.

\bibitem{Noroozi2016}
Mehdi Noroozi and Paolo Favaro.
\newblock Unsupervised learning of visual representations by solving jigsaw
  puzzles.
\newblock In {\em ECCV}, 2016.

\bibitem{Oord2018}
Aaron van~den Oord, Yazhe Li, and Oriol Vinyals.
\newblock Representation learning with contrastive predictive coding.
\newblock {\em arXiv:1807.03748}, 2018.

\bibitem{Oord2017}
Aaron van~den Oord, Oriol Vinyals, and Koray Kavukcuoglu.
\newblock Neural discrete representation learning.
\newblock In {\em NeurIPS}, 2017.

\bibitem{Pathak2017}
Deepak Pathak, Ross Girshick, Piotr Doll{\'a}r, Trevor Darrell, and Bharath
  Hariharan.
\newblock Learning features by watching objects move.
\newblock In {\em CVPR}, 2017.

\bibitem{Pathak2016}
Deepak Pathak, Philipp Krahenbuhl, Jeff Donahue, Trevor Darrell, and Alexei~A
  Efros.
\newblock Context encoders: Feature learning by inpainting.
\newblock In {\em CVPR}, 2016.

\bibitem{Radford2018}
Alec Radford, Karthik Narasimhan, Tim Salimans, and Ilya Sutskever.
\newblock Improving language understanding by generative pre-training.
\newblock 2018.

\bibitem{Radford2019}
Alec Radford, Jeffrey Wu, Rewon Child, David Luan, Dario Amodei, and Ilya
  Sutskever.
\newblock Language models are unsupervised multitask learners.
\newblock 2019.

\bibitem{Raffel2020}
Colin Raffel, Noam Shazeer, Adam Roberts, Katherine Lee, Sharan Narang, Michael
  Matena, Yanqi Zhou, Wei Li, and Peter~J. Liu.
\newblock Exploring the limits of transfer learning with a unified text-to-text
  transformer.
\newblock {\em JMLR}, 2020.

\bibitem{Ramesh2021}
Aditya Ramesh, Mikhail Pavlov, Gabriel Goh, Scott Gray, Chelsea Voss, Alec
  Radford, Mark Chen, and Ilya Sutskever.
\newblock Zero-shot text-to-image generation.
\newblock In {\em ICML}, 2021.

\bibitem{Simonyan2015}
Karen Simonyan and Andrew Zisserman.
\newblock Very deep convolutional networks for large-scale image recognition.
\newblock In {\em ICLR}, 2015.

\bibitem{Szegedy2016a}
Christian Szegedy, Vincent Vanhoucke, Sergey Ioffe, Jonathon Shlens, and
  Zbigniew Wojna.
\newblock Rethinking the inception architecture for computer vision.
\newblock In {\em CVPR}, 2016.

\bibitem{Touvron2021a}
Hugo Touvron, Matthieu Cord, Matthijs Douze, Francisco Massa, Alexandre
  Sablayrolles, and Herv{\'e} J{\'e}gou.
\newblock Training data-efficient image transformers \& distillation through
  attention.
\newblock In {\em ICML}, 2021.

\bibitem{Touvron2021b}
Hugo Touvron, Alexandre Sablayrolles, Matthijs Douze, Matthieu Cord, and
  Herv{\'e} J{\'e}gou.
\newblock Grafit: Learning fine-grained image representations with coarse
  labels.
\newblock In {\em ICCV}, 2021.

\bibitem{Touvron2019}
Hugo Touvron, Andrea Vedaldi, Matthijs Douze, and Herv{\'e} J{\'e}gou.
\newblock Fixing the train-test resolution discrepancy.
\newblock {\em arXiv:1906.06423}, 2019.

\bibitem{VanHorn2018}
Grant Van~Horn, Oisin Mac~Aodha, Yang Song, Yin Cui, Chen Sun, Alex Shepard,
  Hartwig Adam, Pietro Perona, and Serge Belongie.
\newblock The {iNaturalist} species classification and detection dataset.
\newblock In {\em CVPR}, 2018.

\bibitem{Vaswani2017}
Ashish Vaswani, Noam Shazeer, Niki Parmar, Jakob Uszkoreit, Llion Jones,
  Aidan~N Gomez, Lukasz Kaiser, and Illia Polosukhin.
\newblock Attention is all you need.
\newblock In {\em NeurIPS}, 2017.

\bibitem{Vincent2008}
Pascal Vincent, Hugo Larochelle, Yoshua Bengio, and Pierre-Antoine Manzagol.
\newblock Extracting and composing robust features with denoising autoencoders.
\newblock In {\em ICML}, 2008.

\bibitem{Vincent2010}
Pascal Vincent, Hugo Larochelle, Isabelle Lajoie, Yoshua Bengio, Pierre-Antoine
  Manzagol, and L{\'e}on Bottou.
\newblock Stacked denoising autoencoders: Learning useful representations in a
  deep network with a local denoising criterion.
\newblock {\em JMLR}, 2010.

\bibitem{Wang2019}
Haohan Wang, Songwei Ge, Zachary Lipton, and Eric~P Xing.
\newblock Learning robust global representations by penalizing local predictive
  power.
\newblock In {\em NeurIPS}, 2019.

\bibitem{Wang2015a}
Xiaolong Wang and Abhinav Gupta.
\newblock Unsupervised learning of visual representations using videos.
\newblock In {\em ICCV}, 2015.

\bibitem{Wu2018a}
Zhirong Wu, Yuanjun Xiong, Stella Yu, and Dahua Lin.
\newblock Unsupervised feature learning via non-parametric instance
  discrimination.
\newblock In {\em CVPR}, 2018.

\bibitem{Xiao2018}
Tete Xiao, Yingcheng Liu, Bolei Zhou, Yuning Jiang, and Jian Sun.
\newblock Unified perceptual parsing for scene understanding.
\newblock In {\em ECCV}, 2018.

\bibitem{Xiao2021}
Tete Xiao, Mannat Singh, Eric Mintun, Trevor Darrell, Piotr Doll{\'a}r, and
  Ross Girshick.
\newblock Early convolutions help transformers see better.
\newblock In {\em NeurIPS}, 2021.

\bibitem{Yosinski2014}
Jason Yosinski, Jeff Clune, Yoshua Bengio, and Hod Lipson.
\newblock How transferable are features in deep neural networks?
\newblock In {\em NeurIPS}, 2014.

\bibitem{You2017}
Yang You, Igor Gitman, and Boris Ginsburg.
\newblock Large batch training of convolutional networks.
\newblock {\em arXiv:1708.03888}, 2017.

\bibitem{Yuan2021}
Li Yuan, Qibin Hou, Zihang Jiang, Jiashi Feng, and Shuicheng Yan.
\newblock {VOLO}: Vision outlooker for visual recognition.
\newblock {\em arXiv:2106.13112}, 2021.

\bibitem{Yun2019}
Sangdoo Yun, Dongyoon Han, Seong~Joon Oh, Sanghyuk Chun, Junsuk Choe, and
  Youngjoon Yoo.
\newblock Cutmix: Regularization strategy to train strong classifiers with
  localizable features.
\newblock In {\em ICCV}, 2019.

\bibitem{Zhang2018a}
Hongyi Zhang, Moustapha Cisse, Yann~N Dauphin, and David Lopez-Paz.
\newblock mixup: Beyond empirical risk minimization.
\newblock In {\em ICLR}, 2018.

\bibitem{Zhang2016}
Richard Zhang, Phillip Isola, and Alexei~A Efros.
\newblock Colorful image colorization.
\newblock In {\em ECCV}, 2016.

\bibitem{Zhou2014}
Bolei Zhou, Agata Lapedriza, Jianxiong Xiao, Antonio Torralba, and Aude Oliva.
\newblock Learning deep features for scene recognition using {Places} database.
\newblock In {\em NeurIPS}, 2014.

\bibitem{Zhou2019}
Bolei Zhou, Hang Zhao, Xavier Puig, Tete Xiao, Sanja Fidler, Adela Barriuso,
  and Antonio Torralba.
\newblock Semantic understanding of scenes through the {ADE20K} dataset.
\newblock {\em IJCV}, 2019.

\end{thebibliography}

\clearpage
\newpage
\appendix

\section{Implementation Details}\label{app:impl}

\subsection{ImageNet Experiments}\label{app:impl_mae}

\paragraph{ViT architecture.} We follow the standard ViT architecture \cite{Dosovitskiy2021}. It has a stack of Transformer blocks \cite{Vaswani2017}, and each block consists of a multi-head self-attention block and an MLP block, both having LayerNorm (LN) \cite{Ba2016}. The encoder ends with LN. As the MAE encoder and decoder have different width, we adopt a linear projection layer after the encoder to match it. Our MAE adds positional embeddings \cite{Vaswani2017} (the sine-cosine version) to both the encoder and decoder inputs. Our MAE does \textit{not} use relative position or layer scaling (which are used in the code of \cite{Bao2021}).

We extract features from the encoder output for fine-tuning and linear probing. As ViT has a class token \cite{Dosovitskiy2021}, to adapt to this design, in our MAE pre-training we append an auxiliary dummy token to the encoder input. This token will be treated as the class token for training the classifier in linear probing and fine-tuning. Our MAE works similarly well without this token (with average pooling).

\paragraph{Pre-training.} The default setting is in Table~\ref{tab:impl_mae_pretrain}. We do \textit{not} use color jittering, drop path, or gradient clip. We use xavier\_uniform \cite{Glorot2010} to initialize all Transformer blocks, following ViT's official code \cite{Dosovitskiy2021}. We use the linear \textit{lr} scaling rule \cite{Goyal2017}: \textit{lr} = \textit{base\_lr}$\times$batchsize / 256.

\paragraph{End-to-end fine-tuning.} Our fine-tuning follows common practice of supervised ViT training. The default setting is in Table~\ref{tab:impl_mae_finetune}. We use layer-wise \textit{lr} decay \cite{Clark2020} following \cite{Bao2021}.

\paragraph{Linear probing.} Our linear classifier training follows \cite{Chen2021a}. See Table~\ref{tab:impl_mae_linear}. We observe that linear probing requires a very different recipe than end-to-end fine-tuning. In particular, regularization is in general harmful for linear probing. Following \cite{Chen2021a}, we disable many common regularization strategies: we do \textit{not} use mixup \cite{Zhang2018a}, cutmix \cite{Yun2019}, drop path \cite{Huang2016}, or color jittering, and we set weight decay as zero.

It is a common practice to normalize the classifier input when training a classical linear classifier (\eg, SVM \cite{Cortes1995}). Similarly, it is beneficial to normalize the pre-trained features when training the linear probing classifier. Following \cite{Doersch2015}, we adopt an extra BatchNorm layer \cite{Ioffe2015} without affine transformation (\texttt{\small affine=False}). This layer is applied on the pre-trained features produced by the encoder, and is before the linear classifier. We note that the layer does \textit{not} break the linear property, and it can be absorbed into the linear classifier after training: it is essentially a re-parameterized linear classifier.\footnotemark~Introducing this layer helps calibrate the feature magnitudes across different variants in our ablations, so that they can use the same setting without further \textit{lr} search.

\begin{table}[t]
\tablestyle{6pt}{1.02}
\scriptsize
\begin{tabular}{y{96}|y{68}}
config & value \\
\shline
optimizer & AdamW \cite{Loshchilov2019} \\
base learning rate & 1.5e-4 \\
weight decay & 0.05 \\
optimizer momentum & $\beta_1, \beta_2{=}0.9, 0.95$ \cite{Chen2020c} \\
batch size & 4096 \\
learning rate schedule & cosine decay \cite{Loshchilov2016} \\
warmup epochs \cite{Goyal2017} & 40 \\
augmentation & RandomResizedCrop \\
\end{tabular}
\vspace{-.5em}
\caption{\textbf{Pre-training setting.}}
\label{tab:impl_mae_pretrain} \vspace{-.5em}
\end{table}

\begin{table}[t]
\tablestyle{6pt}{1.02}
\scriptsize
\begin{tabular}{y{96}|y{68}}
config & value \\
\shline
optimizer & AdamW \\
base learning rate & 1e-3 \\
weight decay & 0.05 \\
optimizer momentum & $\beta_1, \beta_2{=}0.9, 0.999$ \\
layer-wise lr decay \cite{Clark2020,Bao2021} & 0.75 \\
batch size & 1024 \\
learning rate schedule & cosine decay \\
warmup epochs & 5 \\
training epochs & 100 (B), 50 (L/H) \\
augmentation & RandAug (9, 0.5) \cite{Cubuk2020} \\
label smoothing \cite{Szegedy2016a} & 0.1 \\
mixup \cite{Zhang2018a} & 0.8 \\
cutmix \cite{Yun2019} & 1.0 \\
drop path \cite{Huang2016} & 0.1 (B/L) 0.2 (H) \\
\end{tabular}
\vspace{-.5em}
\caption{\textbf{End-to-end fine-tuning setting.}}
\label{tab:impl_mae_finetune} \vspace{-.5em}
\end{table}

\begin{table}[t]
\tablestyle{6pt}{1.02}
\scriptsize
\begin{tabular}{y{96}|y{68}}
config & value \\
\shline
optimizer & LARS \cite{You2017} \\
base learning rate & 0.1 \\
weight decay & 0 \\
optimizer momentum & 0.9 \\
batch size & 16384 \\
learning rate schedule & cosine decay \\
warmup epochs & 10 \\
training epochs & 90 \\
augmentation & RandomResizedCrop \\
\end{tabular}
\vspace{-.5em}
\caption{\textbf{Linear probing setting.} We use LARS with a large batch for faster training; SGD works similarly with a 4096 batch.
\label{tab:impl_mae_linear}}
\end{table}

\footnotetext{Alternatively, we can pre-compute the mean and std of the features and use the normalized features to train linear classifiers.}

\paragraph{Partial fine-tuning.} Our MAE partial fine-tuning (\S\ref{sec:partial_ft}) follows the setting in Table \ref{tab:impl_mae_finetune}, except that we adjust the number of fine-tuning epochs. We observe that tuning fewer blocks requires a longer schedule. We set the numbers of fine-tuning epochs as \{50, 100, 200\} and use the optimal one for each number of blocks tuned.

\subsection{Supervised Training ViT-L/H from Scratch}
\label{app:supervised_vit_large}

We find that it is nontrivial to train \textit{supervised} \mbox{ViT-L/H} \textit{from scratch} on ImageNet-1K. The training is unstable. While there have been strong baselines with publicly available implementations \cite{Touvron2021a} for smaller models, the recipes for the larger ViT-L/H are unexplored. Directly applying the previous recipes to these larger models does not work. A NaN loss is frequently observed during training.

We provide our recipe in Table~\ref{tab:impl_supervised_vit_large}. We use a \textit{wd} of 0.3, a large batch size of 4096, and a long warmup, following the original ViT \cite{Dosovitskiy2021}. We use $\beta_2{=}0.95$ following \cite{Chen2020c}. We use the regularizations listed in Table~\ref{tab:impl_supervised_vit_large} and disable others, following \cite{Xiao2021}. All these choices are for improving training stability. Our recipe can finish training with no NaN loss. The accuracy is 82.6\% for ViT-L (81.5\% w/o EMA), and 83.1\% for ViT-H (80.9\% w/o EMA). Both ViT-L and \mbox{ViT-H} show an overfitting trend if not using EMA.

As a by-product, our recipe for ViT-B has 82.3\% accuracy (82.1\% w/o EMA), \vs 81.8\% in \cite{Touvron2021a}.

\begin{table}[t]
\tablestyle{6pt}{1.02}
\scriptsize
\begin{tabular}{y{96}|y{68}}
config & value \\
\shline
optimizer & AdamW \\
base learning rate & 1e-4 \\
weight decay & 0.3 \\
optimizer momentum & $\beta_1, \beta_2{=}0.9, 0.95$ \\
batch size & 4096 \\
learning rate schedule & cosine decay \\
warmup epochs & 20 \\
training epochs & 300 (B), 200 (L/H) \\
augmentation & \texttt{RandAug} (9, 0.5) \cite{Cubuk2020} \\
label smoothing \cite{Szegedy2016a} & 0.1 \\
mixup \cite{Zhang2018a} & 0.8 \\
cutmix \cite{Yun2019} & 1.0 \\
drop path \cite{Huang2016} & 0.1 (B), 0.2 (L/H) \\
exp. moving average (EMA) & 0.9999
\end{tabular}
\vspace{-.5em}
\caption{\textbf{Supervised training ViT from scratch.}}
\label{tab:impl_supervised_vit_large}
\vspace{-.5em}
\end{table}

\subsection{Object Detection and Segmentation in COCO}
\label{app:coco}

We adapt the vanilla ViT for the use of an FPN backbone \cite{Lin2017} in Mask R-CNN \cite{He2017}. ViT has a stack of Transformer blocks that all produce feature maps at a single scale (\eg, stride 16). We equally divide this stack into 4 subsets and apply convolutions to upsample or downsample the intermediate feature maps for producing different scales (stride 4, 8, 16, or 32, the same as a standard ResNet \cite{He2016}). FPN is built on these multi-scale maps.

For fair comparisons among different methods, we search for hyper-parameters for each entry in Table~\ref{tab:coco} (including all competitors). The hyper-parameters we search for are the learning rate, weight decay, drop path rate, and fine-tuning epochs. We will release code along with the specific configurations. For full model and training details, plus additional experiments, see~\cite{Li2021}.

\subsection{Semantic Segmentation in ADE20K}
\label{app:ade20k}

We use UperNet \cite{Xiao2018} following the semantic segmentation code of \cite{Bao2021}. We fine-tune end-to-end for 100 epochs with a batch size of 16. We search for the optimal \textit{lr} for each entry in Table~\ref{tab:ade20k} (including all competitors).

The semantic segmentation code of \cite{Bao2021} uses relative position bias \cite{Raffel2020}. Our MAE pre-training does \textit{not} use it. For fair comparison, we turn on relative position bias \textit{only} during transfer learning, initialized as zero. We note that our BEiT reproduction uses relative position bias in \textit{both} pre-training and fine-tuning, following their code.

\subsection{Additional Classification Tasks}
\label{app:class}

We follow the setting in Table~\ref{tab:impl_mae_finetune} for iNaturalist and Places fine-tuning (Table~\ref{tab:cls_transfer}). We adjust the \textit{lr} and fine-tuning epochs for each individual dataset.

\section{Comparison on Linear Probing Results}

\begin{table}[t]
\tablestyle{8pt}{1.05}
\begin{tabular}{clrl}
method & model & params & \multicolumn{1}{c}{acc} \\
\shline
iGPT \cite{Chen2020c} & iGPT-L & 1362 M & 69.0 \\
iGPT \cite{Chen2020c} & iGPT-XL & 6801 M & 72.0 \\
BEiT \cite{Bao2021} & ViT-L & 304 M & 52.1${^\dagger}$ \\
\hline
MAE & ViT-B & 86 M & 68.0 \\
MAE & ViT-L & 304 M & 75.8 \\
MAE & ViT-H & 632 M & 76.6
\end{tabular}
\vspace{-1em}
\caption{\textbf{Linear probing results of masked encoding methods}. Our fine-tuning results are in Table~\ref{tab:imagenet_e2e}. ${^\dagger}$: our implementation.}
\label{tab:imagenet_linear}
\end{table}

In \S\ref{sec:partial_ft} we have shown that linear probing accuracy and fine-tuning accuracy are largely \mbox{\textit{uncorrelated}} and they have different focuses about linear separability. We notice that existing masked image encoding methods are generally less competitive in linear probing (\eg, than contrastive learning). For completeness, in Table~\ref{tab:imagenet_linear} we compare on linear probing accuracy with masking-based methods.

Our MAE with ViT-L has 75.8\% linear probing accuracy. This is substantially better than previous masking-based methods. On the other hand, it still lags behind contrastive methods under this protocol: \eg, MoCo~v3 \cite{Chen2021a} has 77.6\% linear probing accuracy for the ViT-L (Figure~\ref{fig:partial_ft}).

\begin{table}[t]
\tablestyle{2.5pt}{1.02}
\begin{tabular}{l x{24}x{24}x{24}x{24}y{32}}
\multirow{1}{*}{dataset}
& \multicolumn{1}{c}{ViT-B} & \multicolumn{1}{c}{ViT-L} & \multicolumn{1}{c}{ViT-H} & \multicolumn{1}{c}{ViT-H$_\text{448}$}
& \gc{prev best}
\\
\shline
IN-Corruption $\downarrow$ \cite{Hendrycks2019} & 51.7 & 41.8 & \textbf{33.8} & 36.8 & \gc{42.5} \cite{Kim2021} \\
IN-Adversarial \cite{Hendrycks2021} & 35.9 & 57.1 & 68.2 & \textbf{76.7} &  \gc{35.8} \cite{Mao2021} \\
IN-Rendition \cite{Hendrycks2021a} & 48.3 & 59.9 & 64.4 & \textbf{66.5} & \gc{48.7} \cite{Mao2021} \\
IN-Sketch \cite{Wang2019} & 34.5 & 45.3 & 49.6 & \textbf{50.9} & \gc{36.0} \cite{Mao2021} \\
\hline
\multicolumn{3}{l}{\gc{\textit{our supervised training baselines:}}} \\
\gc{IN-Corruption $\downarrow$} & \gc{45.8} & \gc{42.3} & \gc{\textbf{41.3}} \\
\gc{IN-Adversarial} & \gc{27.2} & \gc{29.6} & \gc{\textbf{33.1}} \\
\gc{IN-Rendition} & \gc{49.4} & \gc{\textbf{50.9}} & \gc{50.3} \\
\gc{IN-Sketch} & \gc{35.6} & \gc{37.5} & \gc{\textbf{38.0}}
\end{tabular}
\vspace{-1em}
\caption{\textbf{Robustness evaluation on ImageNet variants} (top-1 accuracy, except for IN-C \cite{Hendrycks2019} which evaluates mean corruption error).
We test the same MAE models (Table~\ref{tab:imagenet_e2e}) on different ImageNet validation sets, \textit{without} any specialized fine-tuning. We provide system-level comparisons with the previous best results.
}
\label{tab:imagenet_robustness}
\vspace{-1em}
\end{table}

\section{Robustness Evaluation on ImageNet}

In Table~\ref{tab:imagenet_robustness} we evaluate the robustness of our models on different variants of ImageNet validation sets. We use the same models fine-tuned on \textit{original} ImageNet (Table~\ref{tab:imagenet_e2e}) and only run inference on the different validation sets, \mbox{\textit{without}} any specialized fine-tuning. 
Table~\ref{tab:imagenet_robustness} shows that our method has strong scaling behavior: increasing the model sizes has significant gains. Increasing the image size helps in all sets but IN-C.
Our results outperform the previous best results (of specialized systems) by large margins.

In contrast, \textit{supervised} training performs much worse (Table~\ref{tab:imagenet_robustness} bottom; models described in \ref{app:supervised_vit_large}). For example, with ViT-H, our MAE pre-training is 35\% better on IN-A (68.2\% vs 33.1\%) than the supervised counterpart.

\newpage

\begin{figure*}[t]\centering\vspace{-3em}
\includegraphics[width=1\linewidth]{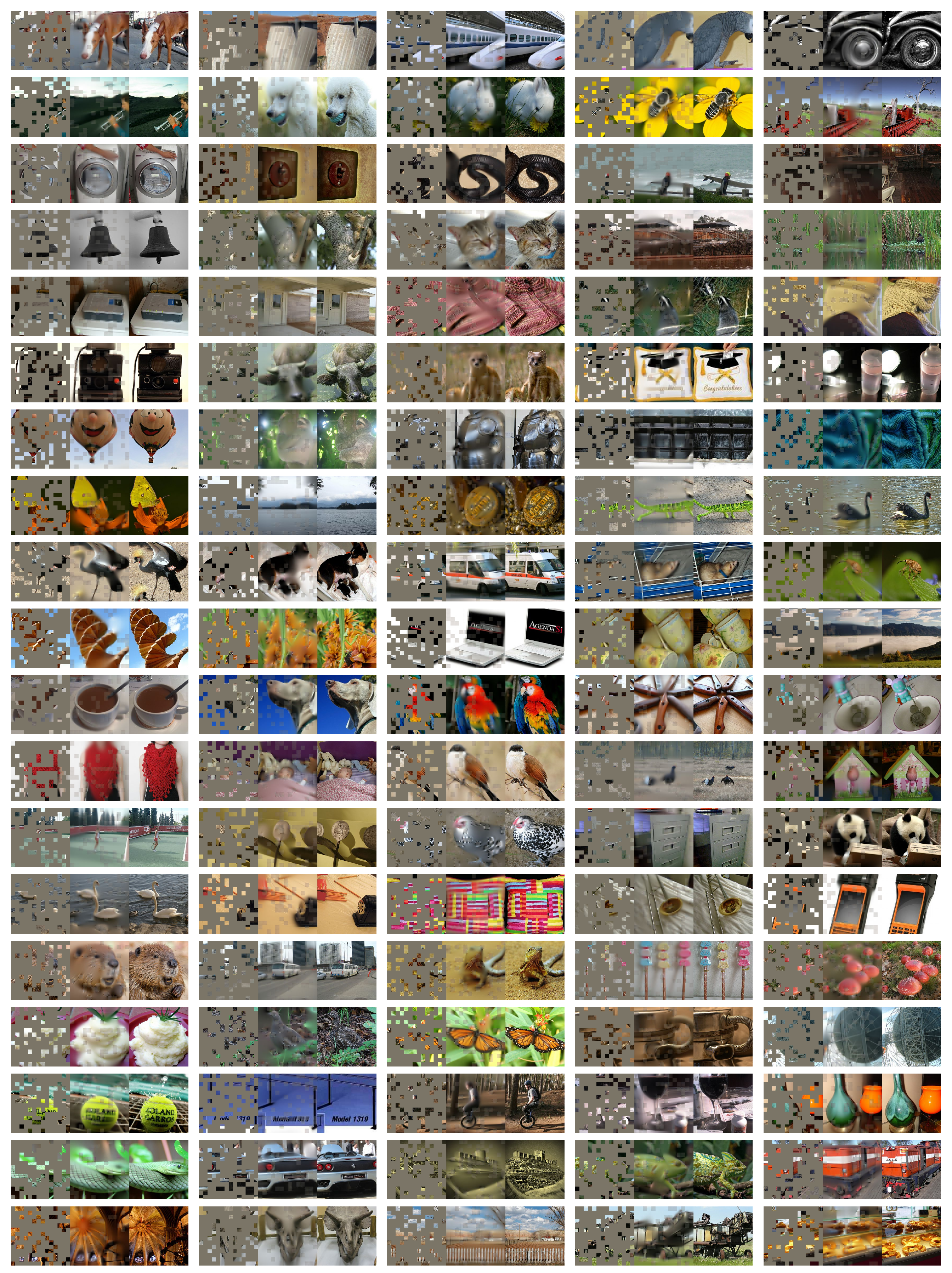}\vspace{-1em}
\caption{\textbf{Uncurated random samples} on ImageNet \textit{validation} images. For each triplet, we show the masked image (left), our MAE reconstruction (middle), and the ground-truth (right). The masking ratio is 75\%.}
\label{fig:samples_uncurated}
\end{figure*}
\vfill

\begin{figure*}[t]\centering\vspace{-3em}
\includegraphics[width=1\linewidth]{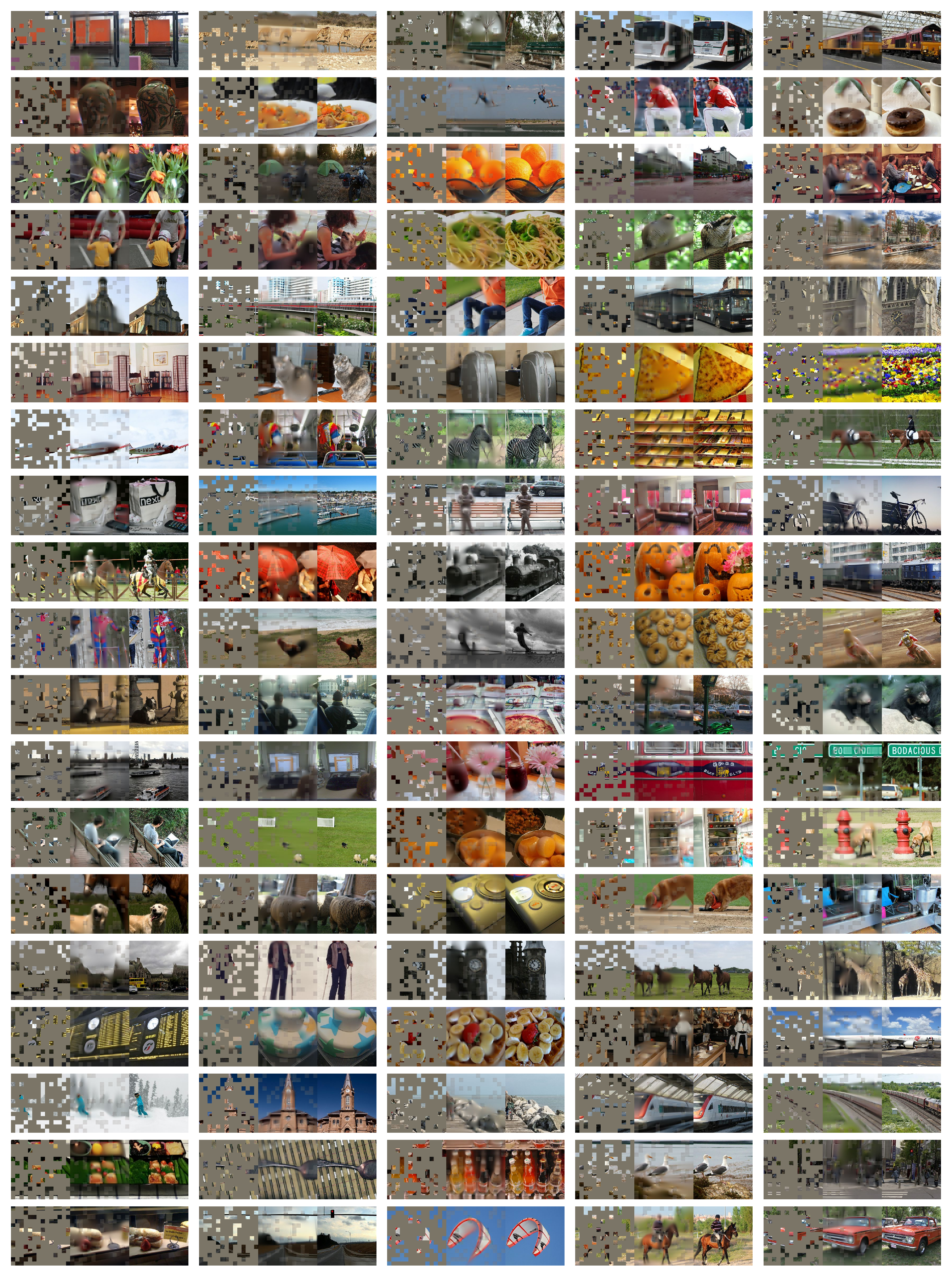}\vspace{-1em}
\caption{\textbf{Uncurated random samples} on COCO validation images, using an MAE trained on ImageNet. For each triplet, we show the masked image (left), our MAE reconstruction (middle), and the ground-truth (right). The masking ratio is 75\%.}
\label{fig:samples_uncurated_coco}
\end{figure*}
\vfill

\end{document}